\documentclass[preprint,review,12pt]{elsarticle}

\usepackage[table,dvipsnames, svgnames, x11names]{xcolor}
\usepackage{graphicx}
\usepackage{amsfonts,amssymb,amsmath,bm,textcomp}
\usepackage{array}
\usepackage[singlelinecheck=false, labelsep=period, font=footnotesize]{caption}
\usepackage[caption=false,font=footnotesize,labelformat=empty,justification=centering]{subfig}
\usepackage[ruled,vlined]{algorithm2e}
\usepackage{mathrsfs,tabularx,multirow,makecell,setspace,subfig,color,booktabs,multirow,algorithmic}
\usepackage{float}
\usepackage{bm}
\usepackage{flushend}
\usepackage[american]{babel}
\usepackage[colorlinks, citecolor=black, bookmarks=true, CJKbookmarks=true, urlcolor=black, bookmarksnumbered=true, bookmarksopen=true]{hyperref}
\usepackage{balance}
\usepackage[utf8]{inputenc}
\usepackage{tikz}
\usepackage{bbm}
\usetikzlibrary{spy}

\newcommand{\squishlist}{
	\begin{list}{$\bullet$}
		{ \setlength{\itemsep}{0pt}
			\setlength{\parsep}{2pt}
			\setlength{\topsep}{2pt}
			\setlength{\partopsep}{0pt}
			\setlength{\leftmargin}{1em}
			\setlength{\labelwidth}{1em}
			\setlength{\labelsep}{0.5em} } }
	\newcommand{\squishend}{
\end{list} }

\newcommand{\mat}[1]{\mbox{\bm{$#1$}}}
\newcommand{\vect}[1]{\mbox{\boldmath{$#1$}}}
\newcommand{\ldbracket}{{[\kern-0.17em[}}
\newcommand{\rdbracket}{{]\kern-0.17em]}}

\usepackage{fancyhdr}

\def\ourmethod{{D2F}\xspace}

\journal{Information Fusion}

\begin{document}

\begin{frontmatter}



\title{Defocus to focus: photo-realistic bokeh rendering by fusing defocus and radiance priors}

\author[label1]{Xianrui~Luo\fnref{label2}}
\ead{xianruiluo@hust.edu.cn}
\author[label1]{Juewen~Peng\fnref{label2}}
\ead{juewenpeng@hust.edu.cn}
\author[label1]{Ke~Xian}
\ead{kexian@hust.edu.cn}
\author[label1]{Zijin~Wu}
\ead{zijinwu@hust.edu.cn}
\author[label1]{Zhiguo~Cao\corref{cor1}}
\ead{zgcao@hust.edu.cn}
\cortext[cor1]{Corresponding author}

\affiliation[label1]{organization={Key Laboratory of Image Processing and Intelligent Control, Ministry of Education; School of Artificial Intelligence and Automation, Huazhong University of Science and Technology},
            city={Wuhan},
            postcode={430074},
            country={PR China}}
            
\fntext[label2]{(Xianrui Luo and Juewen Peng contributed equally to this work.) }

\begin{abstract}
	We consider the problem of realistic bokeh rendering from a single all-in-focus image. 
    Bokeh rendering mimics aesthetic shallow depth-of-field (DoF) in professional photography, but these visual effects generated by existing methods suffer from simple flat background blur and blurred in-focus regions, giving rise to unrealistic rendered results.
	In this work, we argue that realistic bokeh rendering should i) model depth relations and distinguish in-focus regions, ii) sustain sharp in-focus regions, and iii) render physically accurate Circle of Confusion (CoC). 
	To this end, we present a Defocus to Focus (D2F) framework to learn realistic bokeh rendering by fusing defocus priors with the all-in-focus image and by implementing radiance priors in layered fusion. 
    Since no depth map is provided, we introduce defocus hallucination to integrate depth by learning to focus. The predicted defocus map implies the blur amount of bokeh and is used to guide weighted layered rendering. 
    In layered rendering, we fuse images blurred by different kernels based on the defocus map. To increase the reality of the bokeh, we adopt radiance virtualization to simulate scene radiance. The scene radiance used in weighted layered rendering reassigns weights in the soft disk kernel to produce the CoC. 
    To ensure the sharpness of in-focus regions, we propose to fuse upsampled bokeh images and original images. We predict the initial fusion mask from our defocus map and refine the mask with a deep network. We evaluate our model on a large-scale bokeh dataset. 
	Extensive experiments show that our approach is capable of rendering visually pleasing bokeh effects in complex scenes. In particular, our solution receives the runner-up award in the AIM 2020 Rendering Realistic Bokeh Challenge.

\end{abstract}






\end{frontmatter}


\section{Introduction}\label{sec:intro}
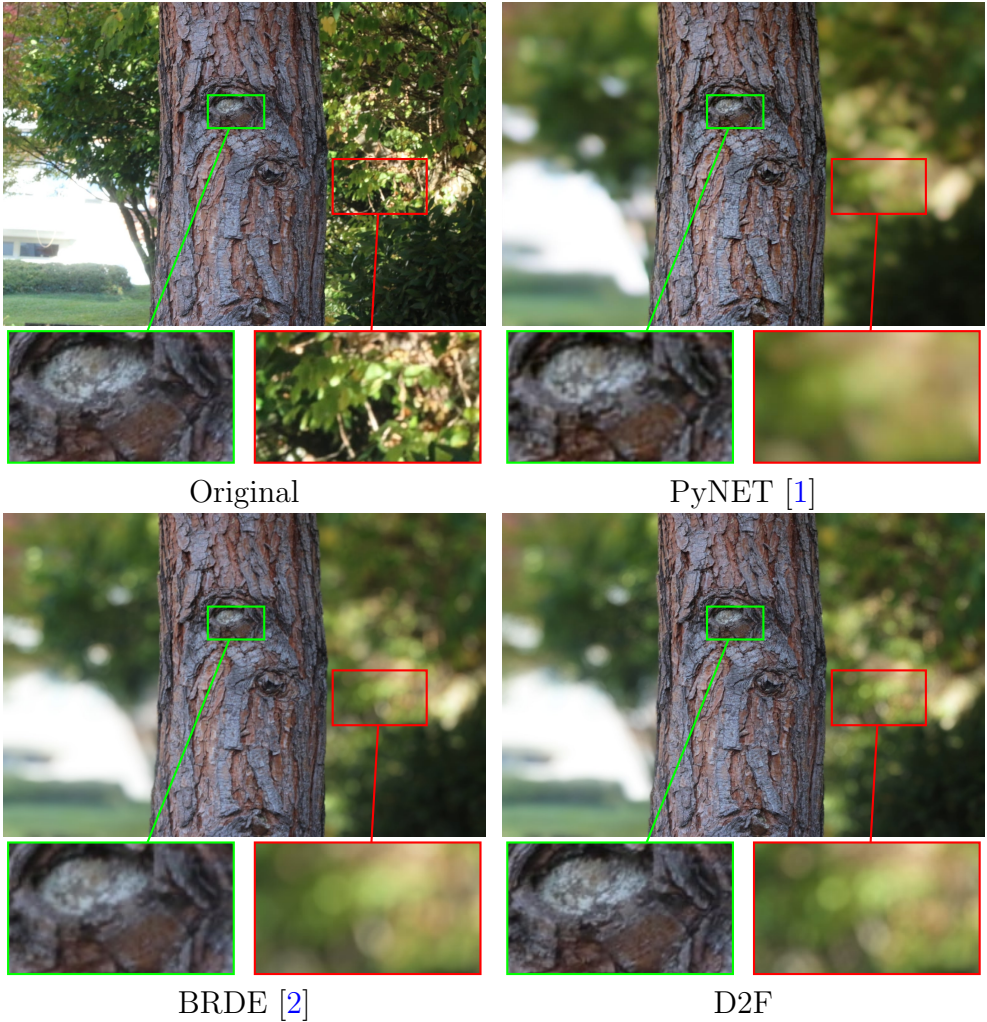
\begin{figure}[!t]
            \centering
            \def\teaserwid{0.467\linewidth}
            \begin{tabular}{*{6}{c@{\hspace{2mm}}}}
                \begin{tikzpicture}[spy using outlines={rectangle, thick,red,magnification=2.4,width=3cm, height=1.75cm,connect spies,spy connection path={\draw[thick] (tikzspyonnode) -- (tikzspyinnode);},every spy on node/.append style={ thick}}]
                \node[inner sep=1]{\includegraphics[width=0.467\linewidth]{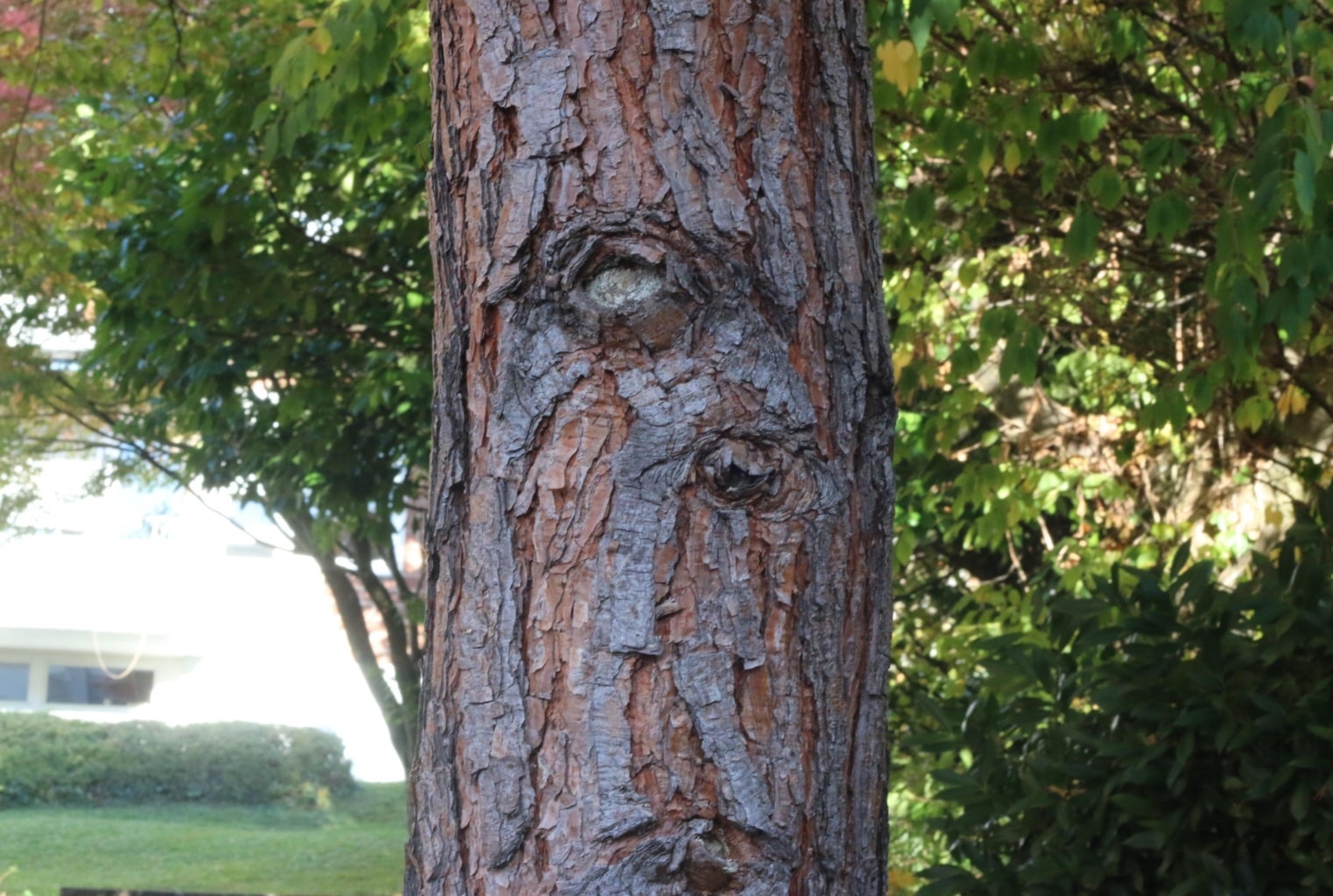}};
                \spy[red] on (1.8,-0.3) in node at (1.64,-3.1);
                \spy[green,magnification=4] on (-0.11,.695) in node at (-1.64,-3.1);
                \end{tikzpicture}& 
                
                \begin{tikzpicture}[spy using outlines={rectangle, thick,red,magnification=2.4,width=3cm, height=1.75cm,connect spies,spy connection path={\draw[thick] (tikzspyonnode) -- (tikzspyinnode);},every spy on node/.append style={ thick}}]
                \node[inner sep=0]{\includegraphics[width=\teaserwid]{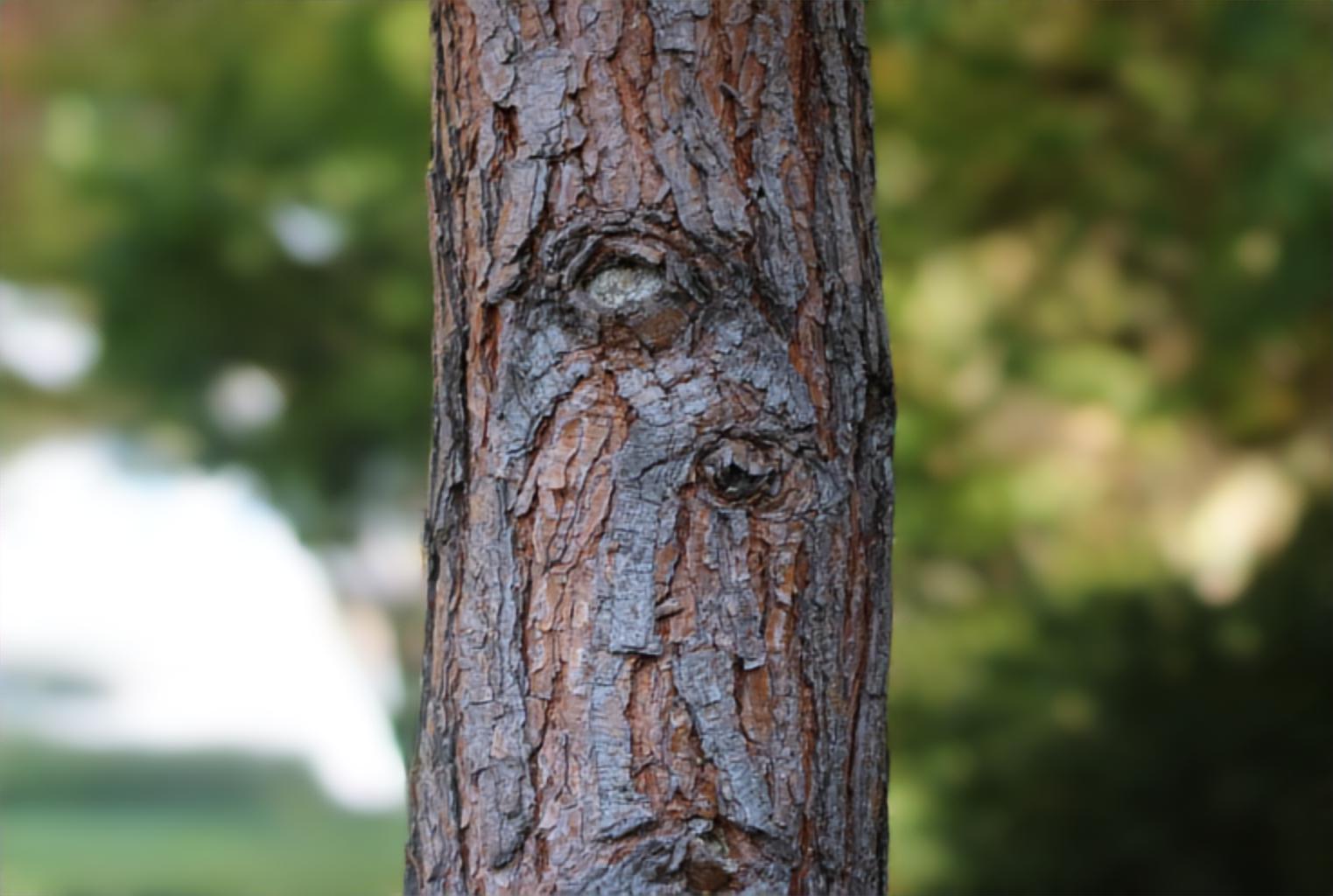}};
                \spy[red] on (1.8,-0.3) in node at (1.64,-3.1);
                \spy[green,magnification=4] on (-0.11,.695) in node at (-1.64,-3.1);
                \end{tikzpicture}& 
                \\
                Original&PyNET~\cite{ignatov2020rendering}\\
                \begin{tikzpicture}[spy using outlines={rectangle, thick,red,magnification=2.4,width=3cm, height=1.75cm,connect spies,spy connection path={\draw[thick] (tikzspyonnode) -- (tikzspyinnode);},every spy on node/.append style={ thick}}]
                \node[inner sep=0]{\includegraphics[width=\teaserwid]{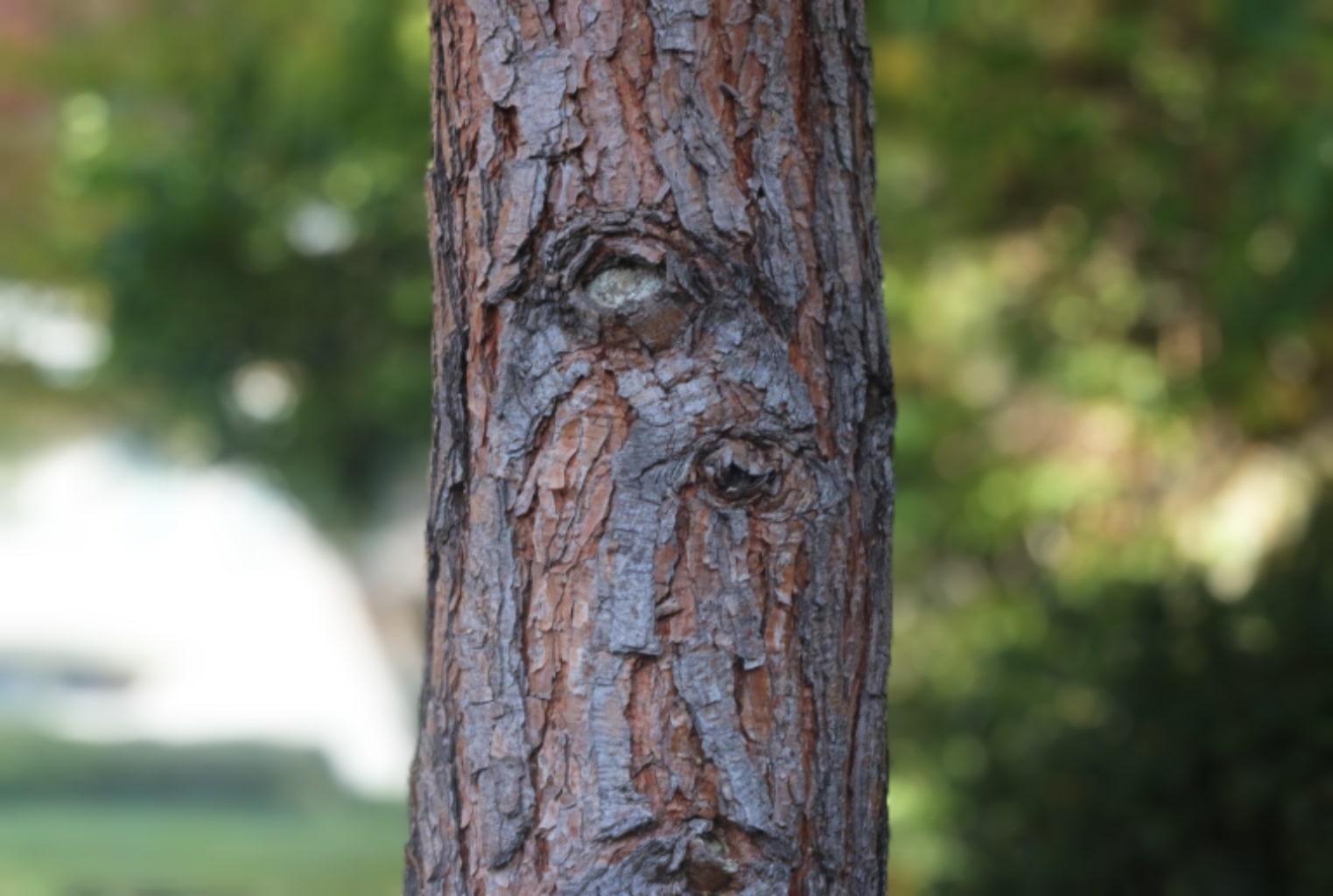}};
                \spy[red] on (1.8,-0.3) in node at (1.64,-3.1);
                \spy[green,magnification=4] on (-0.11,.695) in node at (-1.64,-3.1);
                \end{tikzpicture}& 

                \begin{tikzpicture}[spy using outlines={rectangle, thick,red,magnification=2.4,width=3cm, height=1.75cm,connect spies,spy connection path={\draw[thick] (tikzspyonnode) -- (tikzspyinnode);},every spy on node/.append style={ thick}}]
                \node[inner sep=0]{\includegraphics[width=\teaserwid]{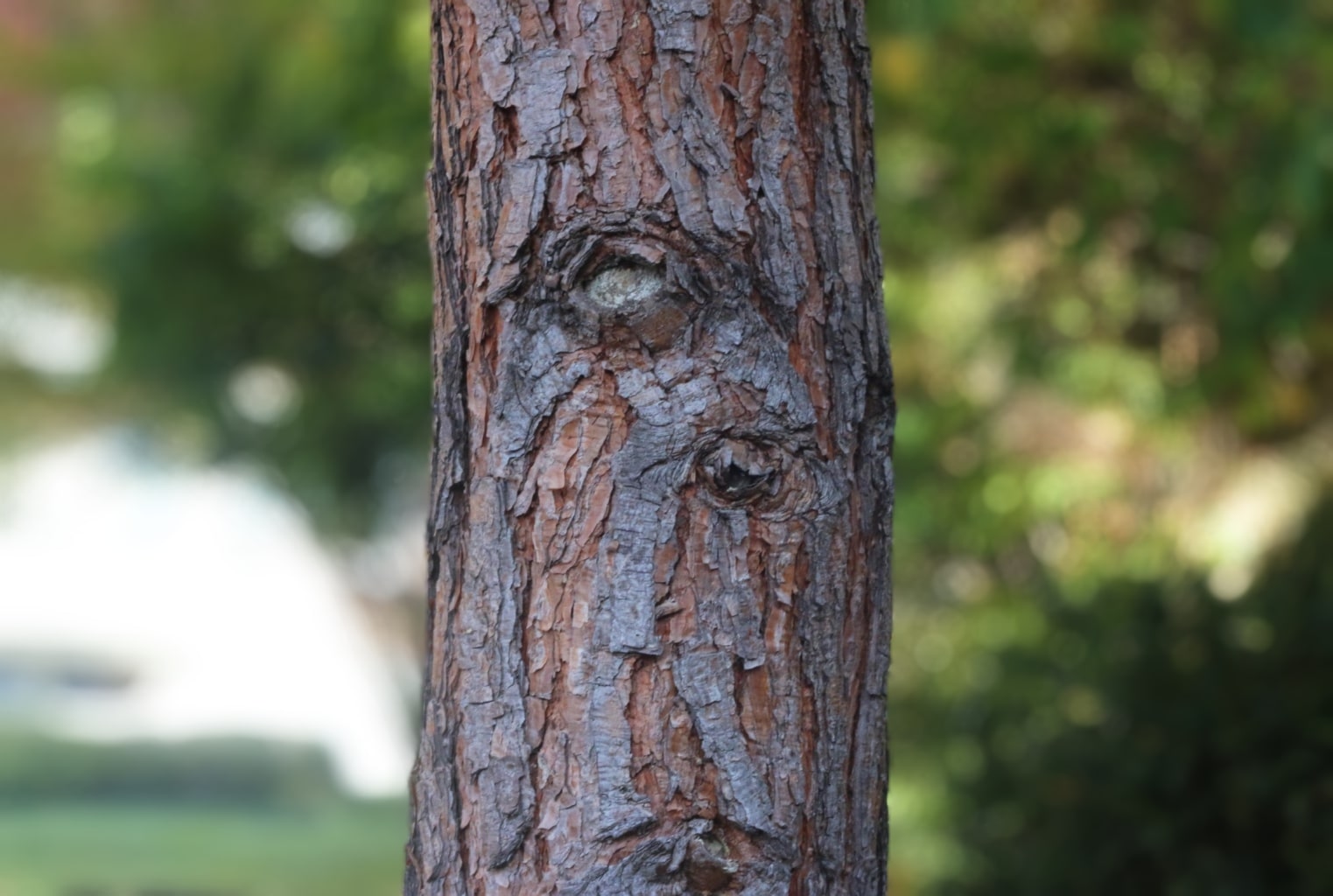}};
                \spy[red] on (1.8,-0.3) in node at (1.64,-3.1);
                \spy[green,magnification=4] on (-0.11,.695) in node at (-1.64,-3.1);
                \end{tikzpicture}& \\
                BRDE~\cite{luo2020bokeh}& D2F\\ 

            \end{tabular}

        \caption{\textbf{
        Comparison of different bokeh rendering approaches}.
        Mainstream bokeh rendering methods 
        such as PyNET~\cite{ignatov2020rendering} cannot explicitly 
        synthesize the Circle of Confusion (CoC) and sometimes can slightly blur in-focus regions. To synthesize the CoC, we present a novel Defocus to Focus (D2F) framework that modulates bokeh with estimated depth and predicted scene radiance.
        During exploration, we find that the blurred in-focus regions are mainly due to the naive treatment of upsampling. We show that this can be addressed in D2F with deep gradient-based image fusion.
        }
        \label{fig:1}
    \end{figure}
	Bokeh, sometimes known as shallow depth of field (DoF), 
	is an important aesthetic feature for photographers, which is popular in 
	videography, portraiture, and landscape photography. 
	Bokeh is closely related to focusing.
	In a camera shot, focusing refers to the process of adjusting the camera lens so that a scene at a certain distance from the camera is clearly imaged.  
	By contrast, generating bokeh is 
	conventionally recognized as blurring out-of-focus regions and producing delightful ``bokeh balls''. The bokeh balls, which are also referred to as the
	Circle of Confusion (CoC), is one of the factors that affect the artistic quality of captured images. The CoC is shaped according to how a lens renders light from out-of-focus areas, and regularly, is viewed as a disc 
	based on the shape of the aperture. 
	In photography, the aperture of a camera is used to control the amount of light that passes through the lens and thus the exposure of the photosensitive sensor; a wide aperture allows more light to travel, which results in shallow DoF. However, not all shallow DoF is equivalent to bokeh. 
	Shallow DoF often blurs the background, but only the blurred background with artistic patterns can be called bokeh.
	
    Bokeh is mostly rendered from an expensive digital single-lens reflex (DSLR) camera by professional photographers, and the settings used to sharpen in-focus areas and to blur 
    the rest require complicated maneuvers. 
    The costs of operation time and expensive hardware hinder easy application of bokeh to amateurs. The desire to easily operate bokeh has therefore motivated enthusiasm for vision-based bokeh rendering. 
	
	Prior work has come up with several ideas to address bokeh rendering. One is to use stereo pairs to obtain disparity maps and to render
	bokeh; however, it requires extra dual-pixel sensors~\cite{wadhwa2018synthetic} or laser scanners~\cite{busam2019sterefo}. 
	Another alternative is deep learning-based neural rendering~\cite{ignatov2020rendering,wang2018deeplens}, where methods are proposed to synthesize blur only from a single all-in-focus image.
	However, most existing models 
	overlook two factors in photo-realistic bokeh rendering: 1) the rendering of CoC, an important artistic feature in bokeh, and 
	2) the sharpness of focused regions. 
	Failing to render them can engender unrealistic effects, as shown in Fig.~\ref{fig:1}.
	
	In this work, we are interested in single-image bokeh rendering. To render photo-realistic bokeh effects, 
	we consider an effective single-image bokeh rendering approach should i)  
	construct correct relative depth relation and perceive in-focus regions, 
	ii) render the physically sound CoC, and iii) maintain the sharpness of focused regions. Indeed rendering requires accurate depth and focusing 
	to imply an appropriate blur amount of each pixel. We aim to integrate these two steps to simplify the pipeline. 
	To address the rendering of CoC, we modify our layered rendering which fuses blurred images with different blur levels. The change is that we reassign weights in rendering, which follows the idea of scene radiance~\cite{yang2016virtual} and manually recovers high dynamic range (HDR)~\cite{Banterle:2017} to synthesize realistic shallow DoF where image intensity is transformed into radiance in HDR.
    In addition, to render the CoC naturally, we design a new disk-like blur kernel. 
	Finally, to keep the sharpness of in-focus regions during upsampling, we propose to execute image fusion beyond naive interpolation. We utilize the predicted defocus map and Poisson gradient constraint as guidance for the fusion mask. 

\begin{figure*}[!tbp]
		\centering
		\includegraphics[width=1.0\linewidth]{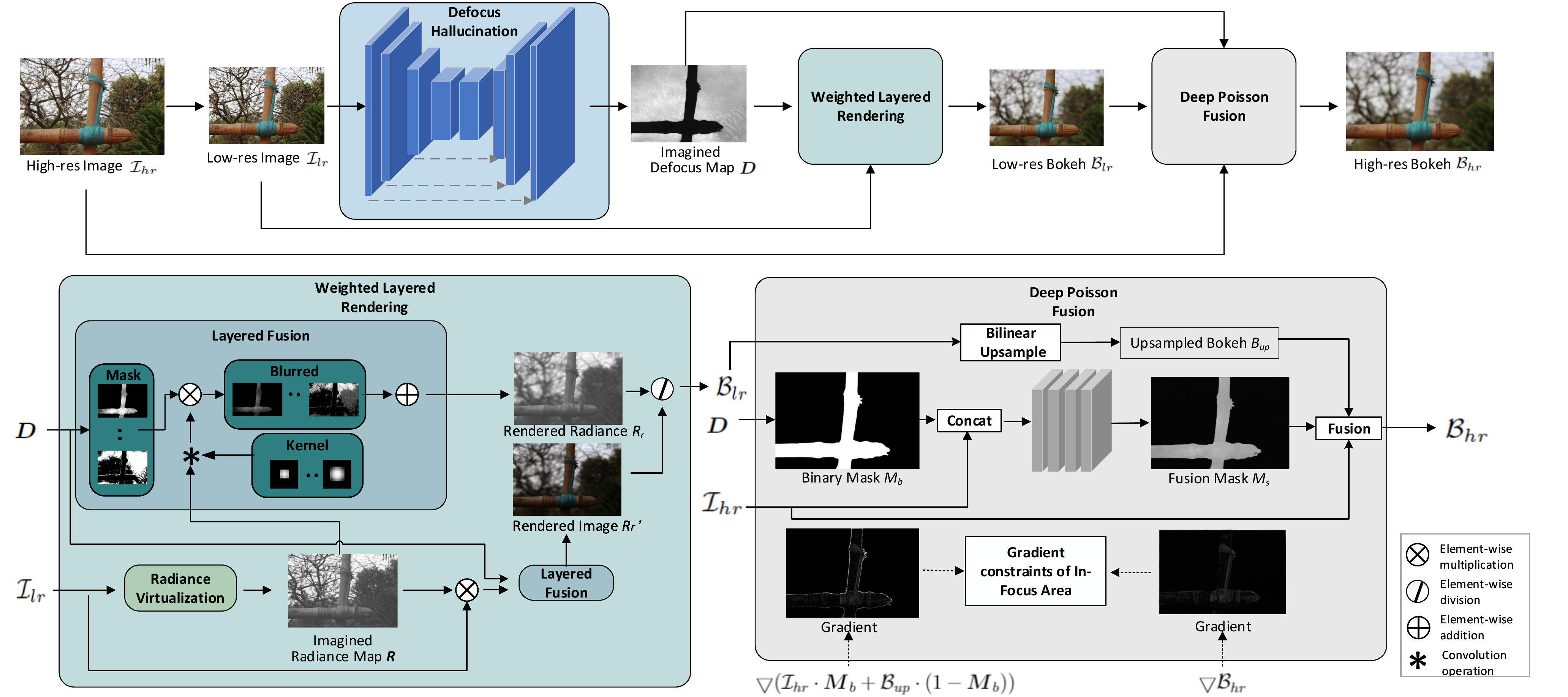}
		\caption{\textbf{\textbf{Technical pipeline of \ourmethod}}. 
		\ourmethod renders photo-realistic bokeh by fusing defocus and radiance priors. 
		In particular,
		a defocus map $\mat{D}$ and a radiance map $\mathcal{R}$ are first predicted with defocus hallucination and radiance virtualization from the low resolution all-in-focus image $\mathcal{I}_{lr}$, respectively, where $\mathcal{I}_{lr}$ is downsampled from its high-resolution counterpart $\mathcal{I}_{hr}$.
        Layered fusion then
        receives two groups of inputs: i) $\mathcal{R}$-weighted $\mathcal{I}_{lr}$ and $\mat{D}$, and ii) $\mathcal{R}$ and $\mat{D}$.
		Next the output $\mathcal{R}_{r}$ and $\mathcal{R}_{r}^{'}$ are used to modulate the low-resolution bokeh $\mathcal{B}_{lr}$. 
        We bilinearly upsample $\mathcal{B}_{lr}$ to generate $\mathcal{B}_{up}$. Finally, we use deep Poisson fusion to generate a mask $\mat{M}_s$ for the fusion of the all-in-focus image $\mathcal{I}_{hr}$ and $\mathcal{B}_{up}$.
        It is worth mentioning that our fusion method produces the mask with the help of $\mat{D}$ and a deep Poisson network.
        This fusion method maintains the sharpness of in-focus area of $\mathcal{B}_{hr}$ during upsampling of $\mathcal{B}_{lr}$.
        }
		\label{fig:pipeline}
	\end{figure*}

	To this end, we present Defocus to Focus (D2F), 
	a fusion framework for photo-realistic bokeh rendering (Fig.~\ref{fig:pipeline}). 
	In particular, we introduce \textit{defocus hallucination} to 
	implement relative depth prediction and focusing. We use the term `hallucination' instead of `estimation' because the predicted defocus map is `imagined' without direct supervision. The imagined defocus map is equivalent to the absolute value of a signed depth map that unifies relative depth and focal distance (`$>0$' indicates the region-to-camera distance is further than focal distance, `$<0$' suggests the distance
	is closer than focal distance, and `$=0$' implies the distance is equal to focal distance). 
    The defocus map is then used as a guidance for \textit{weighted layered rendering}. Compared to layered rendering which fuses images blurred by different blur kernels, we harness scene radiance within pixels and manually set them in HDR to synthesize the CoC. Hence, a \textit{radiance virtualization} module is designed to predict scene radiance, and we choose soft disk blur kernels to increase the reality of CoC.
	For efficiency, defocus hallucination and radiance virtualization are jointly learned to render bokeh in low resolution. 

    To recover the resolution of the rendered result, we propose a novel \textit{deep Poisson fusion} designated for bokeh rendering. We predict the fusion mask by applying the defocus map and training a deep Poisson network. We maintain clear in-focus regions from blurred backgrounds and ensure smooth transitions between them.  
	
	Extensive experiments are conducted to validate \ourmethod quantitatively and qualitatively. We evaluate \ourmethod on the AIM 2020 Rendering Realistic Bokeh Challenge where a large-scale dataset called EBB!~\cite{ignatov2020rendering} is used. 
    {Results show that \ourmethod achieves competitive results against well-established baselines and competition entries. Specifically, \ourmethod increases PSNR by 0.17dB than our previous work~\cite{luo2020bokeh}.}
	To evaluate the performance of the fusion method, we compare our fusion module with other image fusion methods. Our fusion method is proved to outperform other methods.
	We also design ablation studies to prove the effectiveness of each module. 
	Furthermore, we show the superiority of the soft disk blur kernel over the naive one in generating realistic CoC through qualitative visualizations, where the soft blur kernel appears more like a disk. In addition, we compare different training strategies to examine convergence behaviors. 
	Our experiments support that \ourmethod can render photo-realistic bokeh.
	
	Our main contributions include the following:
	\squishlist
	\item D2F: a novel framework which integrates defocus and radiance priors into photo-realistic bokeh rendering. We also apply Poisson fusion to keep the sharpness of in-focus objects;

	\item Defocus Hallucination: a scheme that learns relative depth and focal distance without direct supervision. The defocus map defines the degree of blurring of each pixel, so we can manually change the focal plane as well as the blur amount.

    \item We employ defocus hallucination in deep Poisson fusion, where we predict the fusion mask from the predicted defocus map and Poisson gradient constraint.
	\squishend
	
	The preliminary version of this work appeared in~\cite{luo2020bokeh}, which describes our runner-up solution in the AIM 2020 Rendering Realistic Bokeh Challenge. Here we extend~\cite{luo2020bokeh} in the following aspects. First, we simplify the training scheme for fast convergence.
    Second, we address the remaining issue of our previous pipeline by employing image fusion with the help of a defocus map and a deep Poisson network. Our fusion module can keep the sharpness of in-focus objects.
	Third, we conduct additional experiments and analyses 
	to justify the design choices and the soundness of our inclusions in the D2F framework.

\section{Related Work}\label{relwork}
    \subsection{Defocus Estimation}

    Defocus estimation is closely related to depth estimation~\cite{cao2018monocular,cao2017estimating} and defocus blur detection~\cite{tang2020defusionnet,zhao2021defocus}. While defocus detection focuses on whether the pixel is blurred, the defocus map from defocus estimation represents the amount of defocus blur in shallow DoF images,
    which has many applications such as image deblurring~\cite{zhang2016spatially}, blur magnification~\cite{bae2007defocus}, and depth estimation~\cite{lin2013absolute,Xian_2020_CVPR,shi2015break,kumar2018depth,zhang2013interactive}.
    Defocus estimation can be categorized into two types: region based and edge based. 

    Region based methods 
    directly use image patches to estimate the defocus blur.
    In particular, some works~\cite{shi2015break,shi2015just} 
    focus on detecting small blur amount. Yan {\it et al.}~\cite{yan2016blind} apply a neural network to predict the blur type and its parameters. Tang {\it et al.}~\cite{tang2016spectral} use log averaged spectrum residual to obtain a coarse defocus map and refine it iteratively by exploiting the relevance of similar neighbor regions. 

    In edge based
    paradigms, an early
    attempt~\cite{zhuo2011defocus} predicts the blur amount around edges from the ratio of gradients between original and blurred images, then interpolation from Laplacian matting~\cite{levin2007closed} is applied to 
    recover the defocus map. Xu {\it et al.}~\cite{xu2017estimating} also use Laplacian matting~\cite{levin2007closed}, but estimate a sparse defocus map from metric ranks of local patches in the gradient domain. Kumar {\it et al.}~\cite{kumar2019simultaneous} estimate motion blur and defocus blur jointly. Park {\it et al.}~\cite{park2017unified} introduce various hand-crafted features and deep features to produce defocus map from multi-scale image patches under the guidance of edge-preserved images. However, sometimes the values in homogeneous areas are inconsistent. In addition, the frequency information of image edges are exploited for defocus estimation, using spectrum contrast~\cite{tang2013defocus} or sub-band decomposition~\cite{chakrabarti2010analyzing}. 

    Interestingly, while the concept of the defocus map is closely related to bokeh, most existing methods neglect its applications in bokeh rendering.
    In our work, we show that the defocus map is an essential component for realistic bokeh rendering. It not only helps layered rendering , but also guides the image fusion in our rendering framework.
    
    \subsection{Shallow Depth of Field}

    Shallow DoF can be applied to
    autofocus systems~\cite{fontaine2017survey,herrmann2020learning,xian2021ranking,zhang2019synthetic}, bokeh synthesis from mobile phone camera~\cite{wadhwa2018synthetic,wang2018deeplens}, and even supervision for monocular depth estimation~\cite{srinivasan2018aperture,tang2017depth}. In addition, shallow DoF is implemented in video deblurring from a defocused video pair~\cite{lin2014extracting}. While a depth video and an all-in-focus video are generated from the defocused pair, we render shallow DoF from an all-in-focus image without the guidance of a depth map. 
    
    Among early methods~\cite{yang2016virtual,pharr2016physically,haeberli1990accumulation,lee2010real}, ray tracing methods~\cite{yang2016virtual,pharr2016physically,lee2010real} can precisely reproduce the in-camera physical 
    ray integration.
    However, they are computationally difficult to solve. Alternatively, accumulation
    buffer~\cite{haeberli1990accumulation} is proposed to render a set of sub-apertures and to accumulate each sub-aperture view. 
    Yet, this method is still time-consuming. 

    Large computational cost of physically based methods 
    has prompted a search for a method to 
    generate blur directly in the image domain. 
    Some methods are therefore introduced to allow users to control focus parameters and blur amount~\cite{busam2019sterefo,wang2018deeplens}.
    Wang {\it et al.}~\cite{wang2018deeplens} propose a multi-stage model
    that combines real data and synthetic data to 
    synthesize a shallow DoF effect from a single image.
    Wadhwa {\it et al.}~\cite{wadhwa2018synthetic} focus on the shallow DoF effect with dual-pixel 
    sensors in the mobile phone camera by segmenting portraits/objects and predicting depth.
    In addition, a blending method
    for depth-based bokeh rendering~\cite{busam2019sterefo,zhang2019synthetic,dutta2021depth} is proposed to generate the shallow DoF based on composition of images blurred by different kernels. The blur kernel can be produced by a scatter~\cite{krivanek2003fast} or a cluster~\cite{robison2009image} operation. Moreover, Ignatov {\it et al.}~\cite{ignatov2020rendering} present a large-scale bokeh dataset and propose a multi-level network to gradually 
    refine low-level details. Xiao {\it et al.}~\cite{xiao2018deepfocus} synthesize physically palusible defocus blur for head-mounted displays in real-time. However this method requires a single RGB-D image as input.
    
    To generate realistic bokeh, we apply a layered rendering idea inspired by~\cite{busam2019sterefo}. 
    The rendered results are fused with predicted radiance maps manually set in HDR to attain CoC effects, resulting in better visual quality.
    
    \subsection{Image Fusion/Blending} 
    Image blending is widely used and refers to a certain region of a source image seamlessly replaces a certain area of a target image. As comparison, image fusion refers to extracting and then combining the most significant information from multiple images, and the generated fused single image is  expected to be more informative for following applications, such as multi-focus fusion~\cite{de2013multi,farid2019multi}, visible and infrared image fusion~\cite{ma2016infrared,ma2019fusiongan}, and medical image fusion~\cite{wang2008medical,singh2014fusion}. 
    
    Image blending aims to achieve smooth transition between the source and the target image. Alpha blending~\cite{uyttendaele2001eliminating} is a simple and fast solution, however, it is too naive and blurs details in blending. A multi-scale blending algorithm~\cite{burt1987laplacian} is put forward to blend images on different scales of Laplacian Pyramids. Gradient-based methods~\cite{perez2003poisson,szeliski2011fast,sun2013poisson,wu2019gp} are commonly used. Perez \textit{et al.}~\cite{perez2003poisson} first integrate a Poisson equation in blending. Farbman \textit{et al.}~\cite{farbman2009coordinates} introduce a coordinate-based approach to improve speed by avoiding solving a large linear system. This method is made faster by implementing convolutions~\cite{farbman2011convolution}. A general framework of image fusion is proposed to evaluate performances on multi-scale transform and sparse representation~\cite{liu2015general}. Zhang \textit{et al.}~\cite{zhang2020deep} propose a Poisson blending loss and jointly optimize style as well as content. However this method optimize one image at a time, and the optimization process is time-consuming.
    
    To maintain sharp in-focus areas and seamlessly blend blurred images and original images, we propose a novel image fusion designated for bokeh rendering. We first predict a initial mask from the predicted defocus map, then we implement the Poisson gradient loss in~\cite{zhang2020deep} and integrate it in our deep image blending.  
	
    \section{Defocus to Focus Framework}
    Bokeh rendering conventionally requires three
    components: (i) depth relations, (ii) the focal plane, and (iii) out-of-focus rendering. 
    In this paper, we propose a Defocus to Focus (D2F) fusion framework to implement the three components. 
    In particular, \ourmethod simplifies (i) and (ii) into 
    defocus hallucination, which integrates depth estimation and focal distance detection. The 
    imagined defocus map can be a useful cue implying the blur amount of bokeh. 
    For efficiency consideration, defocus hallucination is executed at half of the original image resolution.
    
    For realistic out-of-focus rendering, we first introduce radiance virtualization into \ourmethod to simulate transformation from image intensity to scene radiance, inspired by the fact that the real optical rendering in the DSLR camera is based on scene radiance~\cite{lin2011revisiting}. The radiance map is manually set in HDR to increase the visibility of CoC. Given the low-resolution defocus map and the radiance map as weight, we next fuse images blurred by different kernels to generate bokeh, where a soft blur kernel helps the generating process achieve the CoC in a layer-wise manner.
    
    Finally, to recover rendered results to the original resolution, we resort to a fusion strategy to ensure clear refocused regions determined by the defocus map and restrained by Poisson gradient loss. We adopt image fusion because we find that in-focus regions tend to be blurred due to naive upsampling. 
    In summary, \ourmethod consists of $3$ modules: defocus hallucination, weighted layered rendering, and deep Poisson fusion. The pipeline is shown in Fig.~\ref{fig:pipeline}.
    In what follows, we explain each module in detail. The notations of all parameters are listed in Table~\ref{tab:variable}. We describe the algorithm in Alg~\ref{alg}.
    
    \begin{table}[!t] \small
        \centering
        \captionsetup{singlelinecheck=false,labelsep=newline,justification=centering}
        \caption{\scshape Notations of Parameters. The Subscript $l$ Means the Layer $l$}
        \label{tab:variable}
        \begin{tabular}{cc}
            \toprule
            Notation & Remark\\
            \hline
            $\mathcal{I}_{hr} $ & All-in-focus RGB image of the original size\\
            $\mathcal{I}_{lr} $&Low-resolution all-in-focus RGB image\\
            \hline
            $\mat{D}_{d}$&Disparity map\\
            $f$& Disparity of the focal plane\\
            $\mat{D} $ & Defocus map \\
            \hline
            $\mathcal{R}_{0} $& Radiance map\\
            $\mat{M}_h $& Mask of pixels in HDR\\
            $\mat{R} $ & Radiance map in HDR \\
            \hline
            
            $r_l$&The blur radius\\
            $\mat{M}_l$& Mask of a defocus layer\\
            $\mathcal{I}_l$& RGB image of a defocus layer\\
            $\mat{K}(r_l)$&The blur kernel of a defocus layer\\
            $\mathcal{B}_l$& Rendered result of a defocus layer\\
            $\mathcal{B}$& Result of a layered rendering pipeline~\cite{busam2019sterefo}\\
            
            $\mathcal{R}_r$& Rendered radiance map \\
            $\mathcal{R}_r^{'}$& \makecell[c]{Rendered multiplication of\\radiance map and RGB image} \\
            $\mat{M}_b $& Initial binary Mask of fusion\\
            $\mathcal{B}_{t} $& Target bokeh of deep Poisson network\\
            $\mathcal{B}_{up} $& Upsampled bokeh from bilinear interpolation\\
            
            $\mathcal{B}_{lr} $& Low-resolution bokeh\\
            \hline
            $\mat{M}_s $& Mask of deep Poisson fusion\\
            
            $\mathcal{B}_{hr} $&Final rendered bokeh\\
            \hline

            $\mat{D}_m$& Multi-channel defocus map\\
            
            \bottomrule
        \end{tabular}
    \end{table}

	\subsection{Defocus Hallucination}\label{Defocus Estimation}
	
	Defocus hallucination estimates a defocus map. We highlight `hallucination' instead of `estimation' because the defocus map is `imagined' conditioned on the blur kernels, rather than learning from supervision. 
	Since the defocus map is another form of the depth map in depth prediction~\cite{srinivasan2018aperture}, any monocular depth estimation network
	can be used in implementation. In particular,
	we use a well-established architecture~\cite{xian2018monocular}. 
	
	In contrast to the previous use of the defocus map~\cite{bae2007defocus,lin2013absolute}, we
	learn the defocus map in an unsupervised manner:
	a bokeh-free RGB image is used as input, and a single-channel defocus map is output by regression. 
	We show that, even without supervision, the defocus map still can be predicted reasonably due to the latter implicit supervision.

    \begin{algorithm}[htb]
	\caption{The algorithm used in \ourmethod.} 
	\label{alg} 
	\begin{algorithmic}[1]
		\REQUIRE All-in-focus RGB image $I_{hr}$, Maximum weight of bright pixels $\alpha$, Degree of difference among pixels $\beta$, Transition smoothness $\gamma$, Depth layers $L$, Blur kernel size function $c$, Defocus hallucination network $f_{d}$, Radiance virtualization network $f_{r}$, Layered rendering function $f_l$, Mask prediction network $f_{p}$, Bilinear resize function $Resize$
		\ENSURE Final rendered bokeh $B_{hr}$ 
		\STATE $I_{lr}=Resize(I_{hr})$
		\STATE $\mat{D}=f_d(\mathcal{I}_{lr})$ 
		\STATE $\mat{R}_0 = f_{r}(\mathcal{I}_{lr})$ 
		\STATE $\mat{M}_h=|\mathcal{I}_{lr}|< 0.99$
		\STATE $\mathcal{R} = \mat{M}_h \cdot \alpha \mathcal{I}_{lr}^\beta +(1-\mat{M}_h) \cdot \mat{R}_0$ 
		\STATE $\mathcal{R}_r^{'}=f_l(\mathcal{R}\cdot \mathcal{I}_{lr},\mat{D})$
		\STATE $\mathcal{R}_r= f_l(\mathcal{R},\mat{D})$
		\STATE $\mathcal{B}_{lr} =\frac{\mathcal{R}_r^{'}}{\mathcal{R}_r}$
		\STATE $\mat{M}_{s}=f_p(\mathcal{I}_{hr},\mat{D})$
		\STATE $\mathcal{B}_{up}=Resize(B_{lr})$
		\STATE $\mathcal{B}_{hr}=\mat{M}_s \cdot\mathcal{I}_{hr} + (1-\mat{M}_s)\cdot\mathcal{B}_{up}$

	\end{algorithmic} 
    \end{algorithm}

    As aforementioned, defocus hallucination integrates monocular depth estimation and focal plane detection. 
    The defocus map 
    $\mat D$ takes the form
	\begin{equation}\label{eq:defocus}
		\mat{D} =  \mat{D}_{d} - b(f)\,,
	\end{equation}
    where $\mat{D}_{d}$ is the disparity map, $f$ denotes the disparity of the focal plane, and $b(\cdot)$ is a broadcasting function that expands the singleton. In following equations, we neglect $b(\cdot)$ to simplify the notation. 

    Formally, defocus hallucination defines a function $f_d$ such that
    \begin{equation}
        \begin{aligned}           \mat{D}=f_d(\mathcal{I}_{lr})\,,
        \end{aligned}
    \end{equation}
    where 
    $\mathcal{I}_{lr} \in \mathbb{R}^{\frac{H}{2}\times \frac{W}{2}\times3}$
    is the low-resolution all-in-focus RGB image. $H$ and $W$ are the height and width of the 
    high-resolution input $\mathcal I_{hr}$.
    With the intermediate defocus map, we 
	integrate it with the following radiance virtualization and layered rendering to simulate physically based bokeh rendering. 
	Furthermore, defocus hallucination is the guidance of deep Poisson fusion for generation of the fusion mask.
	
    \subsection{Weighted Layered Rendering}\label{sec:bokeh_rendering}
	\subsubsection{Radiance Virtualization}\label{subsubsec:weighted layered rendering}
    
    In real photography, shallow DoF is generated from scene radiance rather than image intensity~\cite{yang2016virtual}. Therefore, we introduce radiance virtualization to transform raw RGB values into radiance. In addition, CoC is an important feature in realistic bokeh rendering and is closely related to radiance in HDR. To generate the CoC in color rendering, digital cameras typically have two rendering pipelines: i) the \textit{photofinishing} model that adopts different imaging modes, and
    ii) the \textit{slide or photographic reproduction} model that uses fixed color rendering~\cite{holm2002color}. 
    Here we suppose that all images are captured in the latter mode and that the radiance of each pixel only depends on its RGB values.
    We apply layered rendering on scene radiance instead of image intensity, so we assign pixels with different weights according to their RGB values to establish the transformation from RGB values to scene radiance.
    As shown in Fig.~\ref{fig:pipeline}, we use a network to compute scene radiance from RGB values and to predict the radiance map as
    \begin{equation}\label{radiance}
        \mat{R}_0 = f_{r}(\mathcal{I}_{lr})\,,
    \end{equation}
    where $f_{r}$ is a network consisting of four convolution layers. The network outputs $\mat{R}_0\in \mathbb{R}^{\frac{H}{2}\times \frac{W}{2}}$, which is considered an `imagined' radiance map. 

    In addition, we observe that some bright pixels whose R, G, or B value is close to the upper bound, i.e., $255$. These pixels are supposed to have larger weights indicating their actual energies than other pixels. A straightforward solution is to use the HDR operator. 
    With HDR, the CoC effect can be enhanced in theory. However, in practice we find that designing a network to perform HDR is not needed, for a simple extension in the range of the radiance map is sufficient.
    Therefore, we deal with these bright pixels separately and allocate large weights to them as
    \begin{equation}\label{eq:hdr}
        \mathcal{R} = \mat{M}_h \cdot \alpha \mathcal{I}_{lr}^\beta +(1-\mat{M}_h) \cdot \mat{R}_0\,,
    \end{equation}
    where $\cdot$ is the element-wise product, $\mathcal{R}\in \mathbb{R}^{\frac{H}{2}\times \frac{W}{2}\times3}$ is the modulated radiance map, and $\mat{M}_h\in \mathbb{R}^{\frac{H}{2}\times \frac{W}{2}}$ is a mask which denotes whether R, G, or B value of each pixel is more than a threshold. $\alpha$ and $\beta$ are two hyperparameters, where $\alpha$ controls the maximum weight of bright pixels, and $\beta$ adjusts the degree of difference among RGB color channels. From experiments, one can see that this operation leads to a similar HDR effect. We remark that radiance virtualization is essential:
    empirically training defocus hallucination alone without radiance virtualization
    can lead to non-convergence of the network sometimes. A plausible explanation is that the radiance map also provides some form of supervision to defocus hallucination.
    
    \subsubsection{Layered Rendering}\label{subsubsec:bokeh_rendering}

    The physically motivated refocusing pipeline proposed in SteReFo~\cite{busam2019sterefo} can mimic the real rendering process efficiently. The core idea is to decompose the scene into different depth layers and fuses them after blurring each layer with pre-defined blur kernels conditioned on a known focus plane.
    
    Given a defocus map $\mat{D}\in \mathbb{R}^{\frac{H}{2}\times \frac{W}{2}}$, 
    we quantize the depth values into $L$ depth layers. Thus, the blur radius of the $l$-th layer $r_l\in\vect{r}$, $\vect{r}\in \mathbb{R}^{L}$, $l=1,...,L$, can be computed by
	\begin{equation} \label{eq:radius}
        r_l = c(l) \,,
	\end{equation}
    where $c(\cdot)$ is a camera-dependent function which allocates the corresponding blur kernel size w.r.t.\ $l$.
    In \ourmethod we define $c(\cdot)$ as a piecewise linear function, because the depth of each layer is considered uneven. From the defocus map $\mat D$, the $l$-th layer of the image $\mathcal{I}_{lr}$ is generated by
    \begin{equation} \label{eq:layered mask}
        \mathcal{I}_l = \mat{M}_l\cdot\mathcal{I}_{lr}\,,
    \end{equation}
    where $\mat{M}_l=\mathbbm{1}(|\mat D-\frac{l}{L}| < \frac{1}{L})$. In implementation the indicator function $\mathbbm{1}(\cdot)$ is replaced with a smooth function according to~\cite{busam2019sterefo} as
    \begin{equation}\label{eq:smooth}
        \mat{M}_l = \frac{1}{2} + \frac{1}{2} \tanh \bigg(\gamma(\frac{1}{L}-|\mat D-\frac{l}{L}|)\bigg)\,.
    \end{equation}
    The value of $\gamma$ defines the smoothness of transition between layers. On the one hand, a smaller value tends to highlight the gap between each layer and make the final fused result more fragmented. On the other hand, a larger value would fade the CoC.
    Thus $\mathcal I_l\in \mathbb{R}^{\frac{H}{2}\times \frac{W}{2}\times3}$ is blurred as
    \begin{equation} \label{eq:layeredbluring}
        \mathcal B_l = \mat{K}(r_l)\ast \mathcal I_l\,,
    \end{equation}
    where $\mathcal B_l\in \mathbb{R}^{\frac{H}{2}\times \frac{W}{2}\times3}$ is the bokeh result of layer $l$, and $\mat{K}(r_l)\in \mathbb{R}^{r_l\times r_l}$ is the blur kernel conditioned on the kernel size $r_l$. 
    Given the rendered $\mathcal{B}_l$'s, fusing them back takes the form
    \begin{equation} \label{eq:composite_0}
        \mathcal{B} = \frac{\sum_{l=1}^{L}\mathcal{B}_{l}\prod_{j=l+1}^{L} (1-\mat{M}_{j})}{\sum_{l=1}^{L}(\mat{K}(r_{l})\ast\mat{M}_{l})\prod_{j=l+1}^{L} (1-\mat{M}_{j})}\,,  
    \end{equation}
    where 
    $\mat M_{L+1}$ is a zero matrix.
    $l=1$ is considered no blur in Eq.~\eqref{eq:layeredbluring}, so $\mathcal{B}_1=\mathcal{I}_1$.
    
    We combine Eq.~\eqref{eq:layered mask}~\eqref{eq:smooth}~\eqref{eq:layeredbluring}~\eqref{eq:composite_0} to define layered rendering as function $f_l$ such that
    \begin{equation} \label{eq:composite}
        \mathcal B=f_l(\mathcal{B}_{l},\mat{M}_l)=f_l(\mathcal{I}_{lr},\mat D)\,, \\
    \end{equation}
    
\begin{figure}
    \centering
	  \subfloat[$k_{h}\left ( x,y,5 \right)\qquad\quad k_{s}\left ( x,y,5 \right)$]{
       \includegraphics[width=0.2\linewidth]{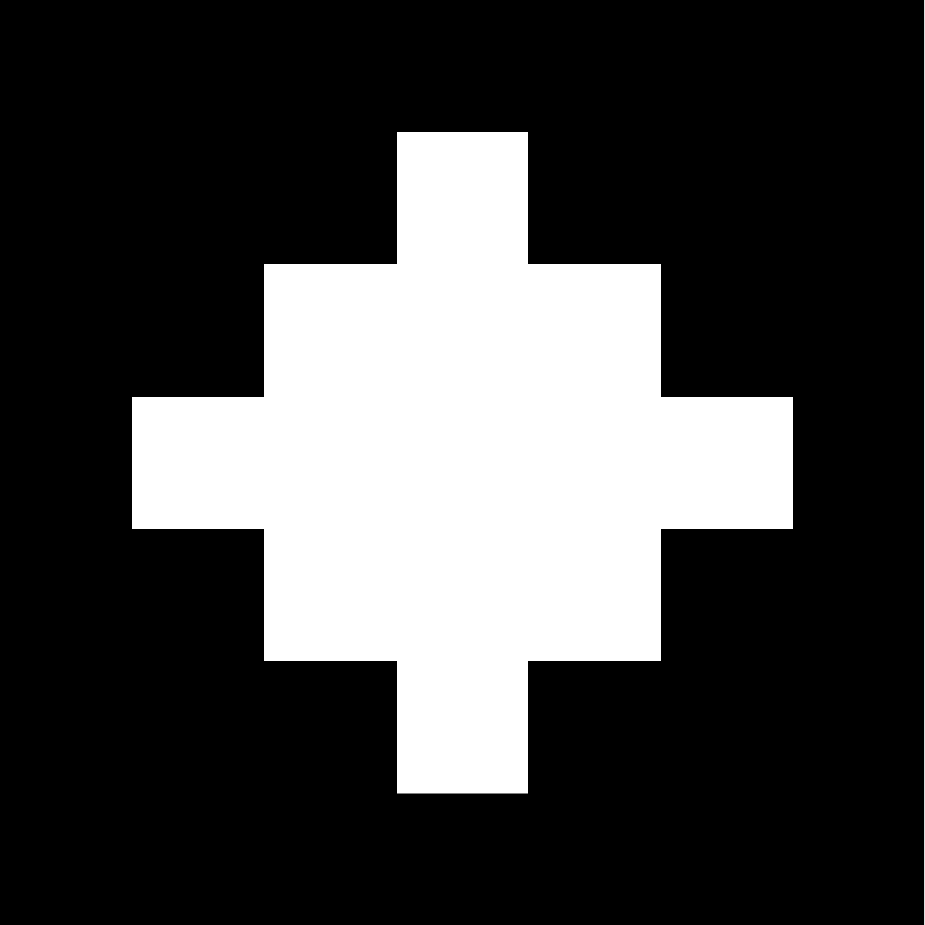}
       \includegraphics[width=0.2\linewidth]{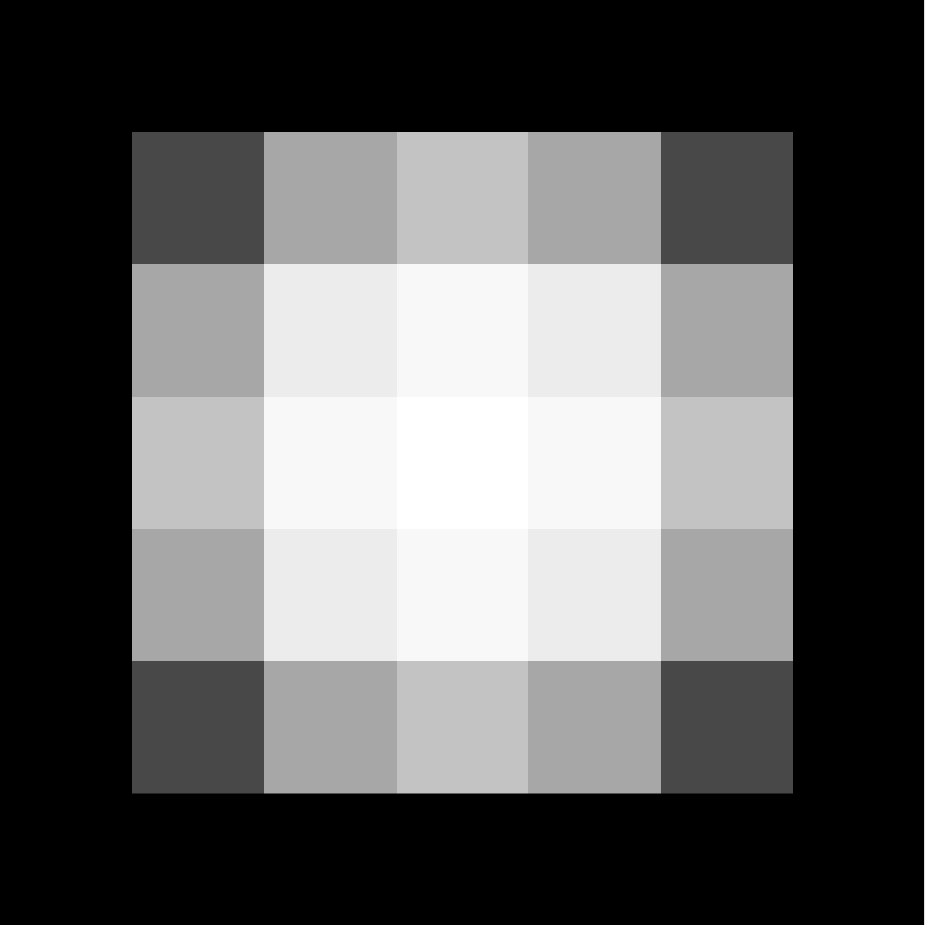}}
    \label{1a}\hfill
	  \subfloat[$k_{h}\left ( x,y,9\right)\qquad\quad k_{s}\left ( x,y,9\right)$]
	  {
        \includegraphics[width=0.2\linewidth]{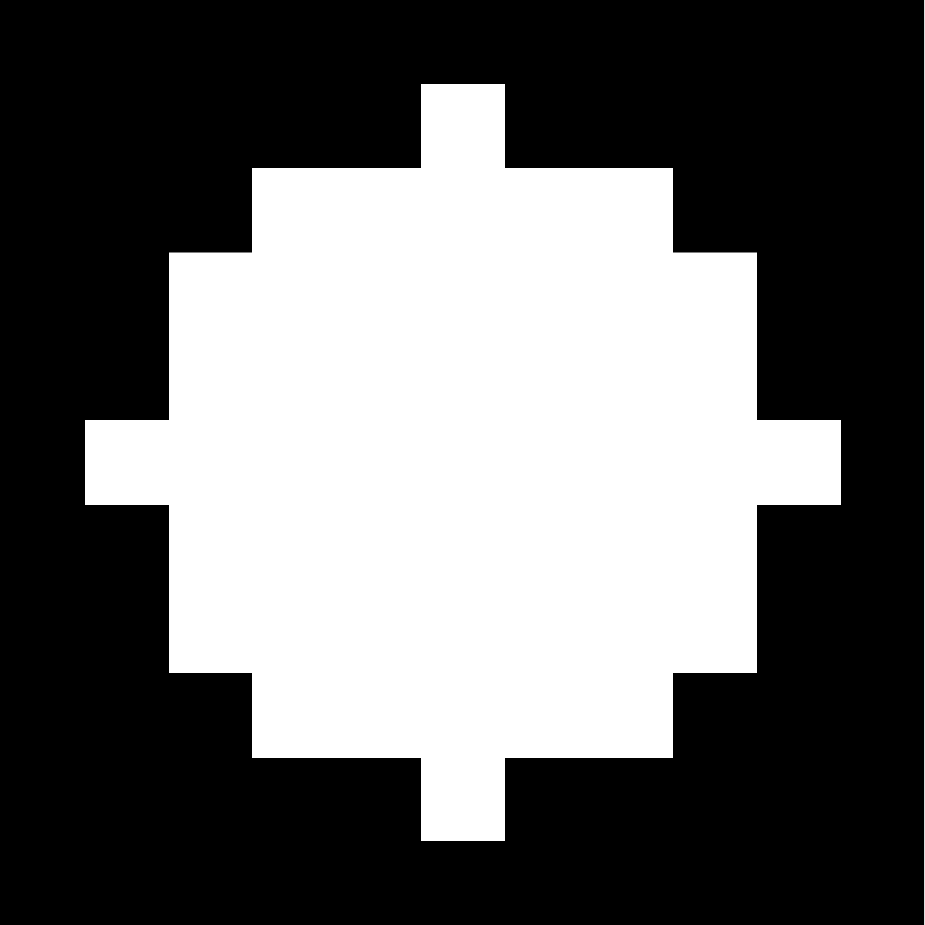}
        \includegraphics[width=0.2\linewidth]{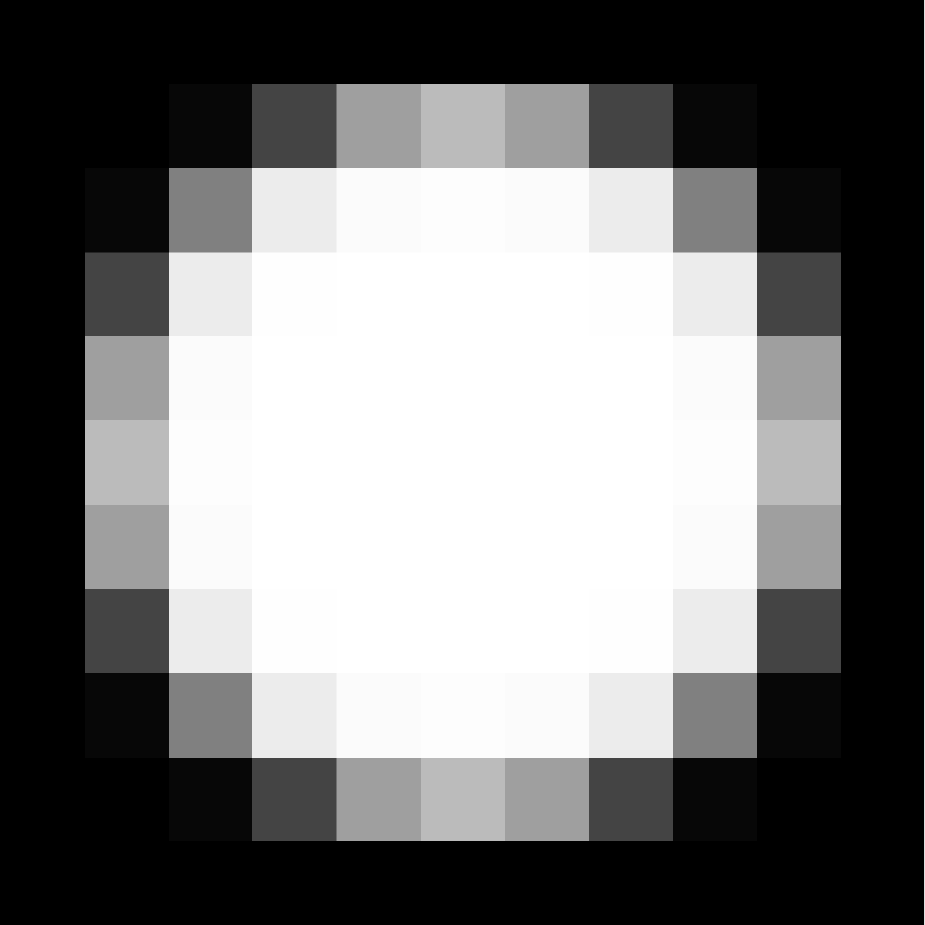}}
    \label{1b}\\

	  \subfloat[$k_{h}\left ( x,y,17 \right)\qquad\quad k_{s}\left ( x,y,17\right)$]
	  {
        \includegraphics[width=0.2\linewidth]{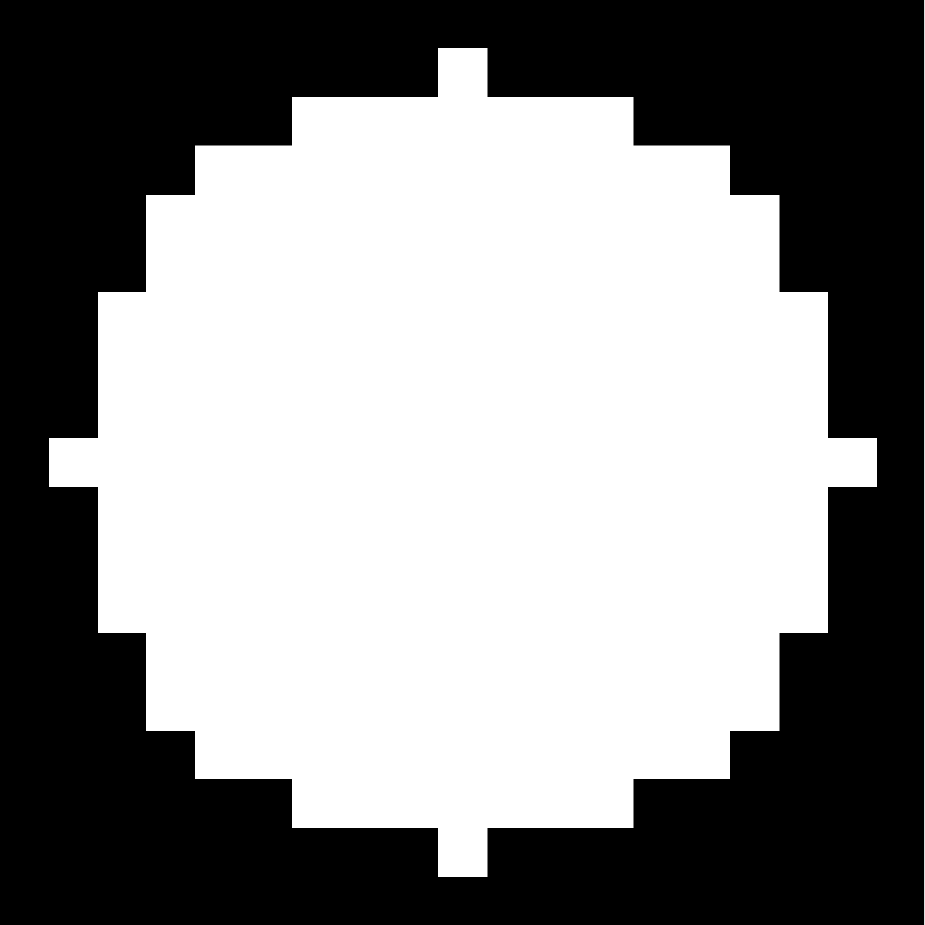}
        \includegraphics[width=0.2\linewidth]{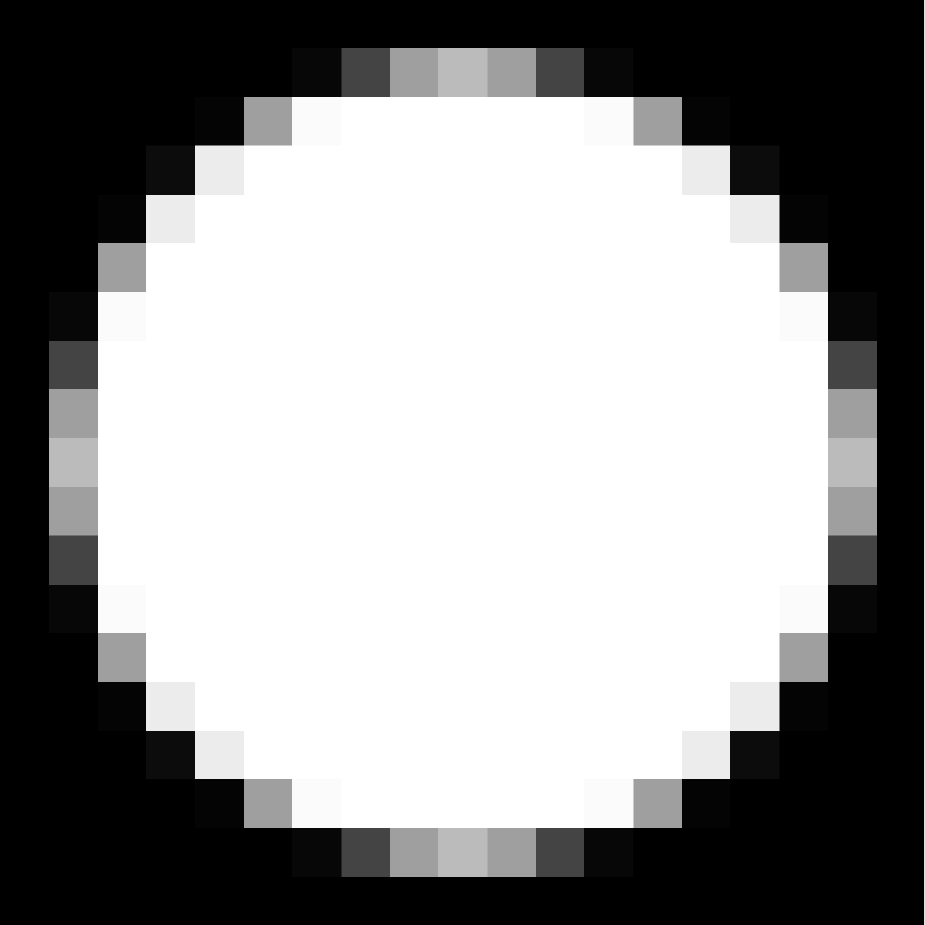}}
    \label{1c}\hfill
	  \subfloat[$k_{h}\left ( x,y,25 \right)\qquad\quad k_{s}\left ( x,y,25\right)$]
	  {
        \includegraphics[width=0.2\linewidth]{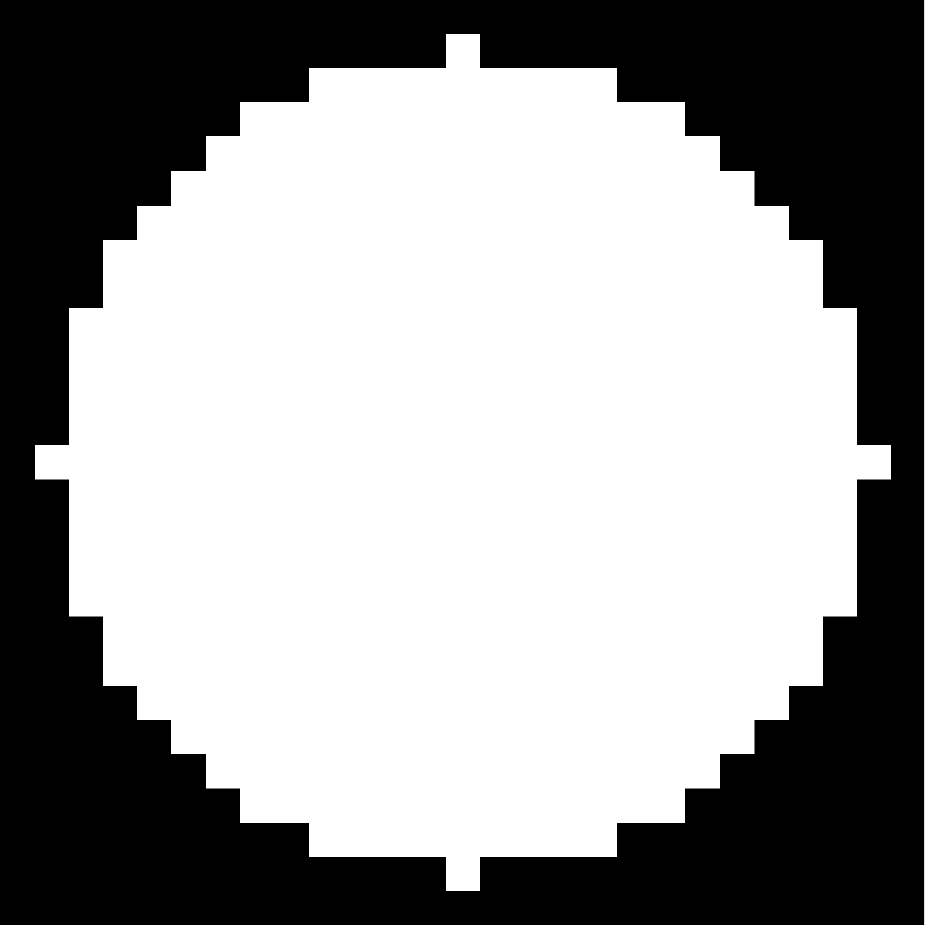}
        \includegraphics[width=0.2\linewidth]{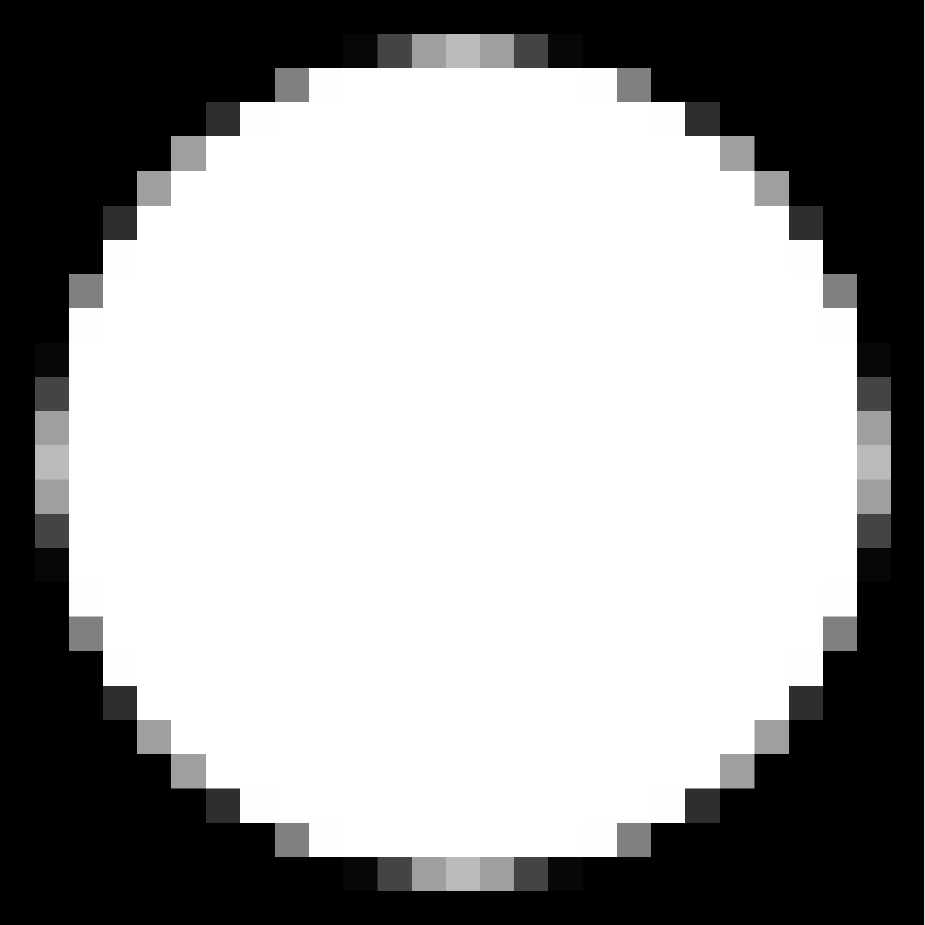}}
    \label{1d} 
    \caption{Visualizations of hard (left) and soft (right) disk blur kernels with different kernel sizes. Visualizations of hard disk blur kernels $k_{h}\left ( x,y,r_l \right)$ and soft disk blur kernels $k_{s}\left ( x,y,r_l \right)$ with different kernel sizes.} 
    \label{fig:kernels}
    \end{figure}

    By integrating layered rendering with defocus hallucination and radiance virtualization, one can attain visually pleasing CoC effects.
    Concretely, our design choices are:
    \squishlist
        \item[i)] \textit{Generating soft blur kernels.} To generate the realistic disk-like CoC, we design a soft disk blur kernel 
        instead of a hard one 
    	formed by $0$ or $1$. For a blur kernel $\mat{K}(r_l)$, the value of a
    	hard kernel follows
	    \begin{equation}
	        k_{h}\left ( x,y,r_l \right)=\left\{\begin{matrix}
	    0, & \text{if}~~r_l^{2}-x^{2}-y^{2}<0 \\ 
	    1, &  \text{otherwise}
	    \end{matrix}\right.\,,
	    \end{equation}
	    where $x$ and $y$ indicate the horizontal and vertical distance relative to the center of the blur kernel. 
	    Note that, this function is not differentiable, a smooth approximation~\cite{busam2019sterefo} can replace as
	    \begin{equation}\label{eq:smooth_hard_kernel}
	        h(z) = \frac{1}{2} + \frac{1}{2} \tanh (z)\,.
	    \end{equation}
	    Following Eq.~\eqref{eq:smooth_hard_kernel}, we define the value in a soft disk kernel by
	    \begin{equation}\label{eq:softkernel}
	        k_s(x,y,r_l) = \frac{1}{2} + \frac{1}{2} \tanh \bigg(\sigma(r_l^2-x^2-y^2)+\varphi\bigg)\,,
	    \end{equation}
	    where $\sigma$ and $\varphi$ control the shape of the disk kernel. The comparison of hard and soft kernels with different sizes is shown in Fig.~\ref{fig:kernels}. One can see that the soft kernel looks more like a disk than the hard
	    one, especially when the kernel size is small. This creates a natural CoC effect. Therefore, \ourmethod uses $k_s(\cdot,\cdot,\cdot)$ to implement $\mat{K}(r_l)$.
    
    \item[ii)] \textit{Customizing blur kernel sizes.} We find that the blur amount of the bokeh image is unbalanced during weighted layered rendering, where a small blur amount tends to appear for most layers in a defocus map. Therefore, we set the intervals of pre-defined kernel sizes 
    in a non-uniform manner. Specifically, we define that $c(\cdot)$ as a piecewise linear function in Eq.~\eqref{eq:radius} such that
    \begin{equation}\label{eq:15kernel}
    c(l)=\left\{
    \begin{array}{rcl}
    2*l+1, & & {1 \leq l<3}\\
    4*(l-3)+7, & & {3 \leq l < 8}\\
    6*(l-8)+27, & & {8 \leq l < 11}\\
    8*(l-11)+45, & & {11 \leq l<15}
    \end{array} \right.\,.
    \end{equation}
    $l=1$ can be considered no blur.
    
    \item[iii)] \textit{Modulating blur with the defocus map and the radiance map.} 
    In \ourmethod we propose to use the predicted defocus map $\mat{D}$ and the radiance map $\mathcal R$ to further modulate blur and to increase the reality of bokeh. We render the radiance map and multiplication of the radiance map and the RGB image separately under the guidance of $\mat{D}$, which outputs $\mathcal{R}_r$ and $\mathcal{R}_r^{'}$. $\mathcal{R}_r^{'}$ divided by $\mathcal{R}_r$ is the rendered bokeh. Such modulation slightly modifies Eq.~\eqref{eq:composite} to
    \begin{equation}
        \mathcal{B}_{lr} =\frac{\mathcal{R}_r^{'}}{\mathcal{R}_r}\,,\\
    \end{equation}
    where 
    \begin{equation}\label{eq:split}
        \begin{split}
        \mathcal{R}_r^{'}=f_l(\mathcal{R}\cdot \mathcal{I}_{lr},\mat{D})\\ 
        \mathcal{R}_r= f_l(\mathcal{R},\mat{D})
        \end{split}\,.
    \end{equation}
    
\begin{figure}[!tbp]
		\centering
		
		\includegraphics[width=1.0\linewidth]{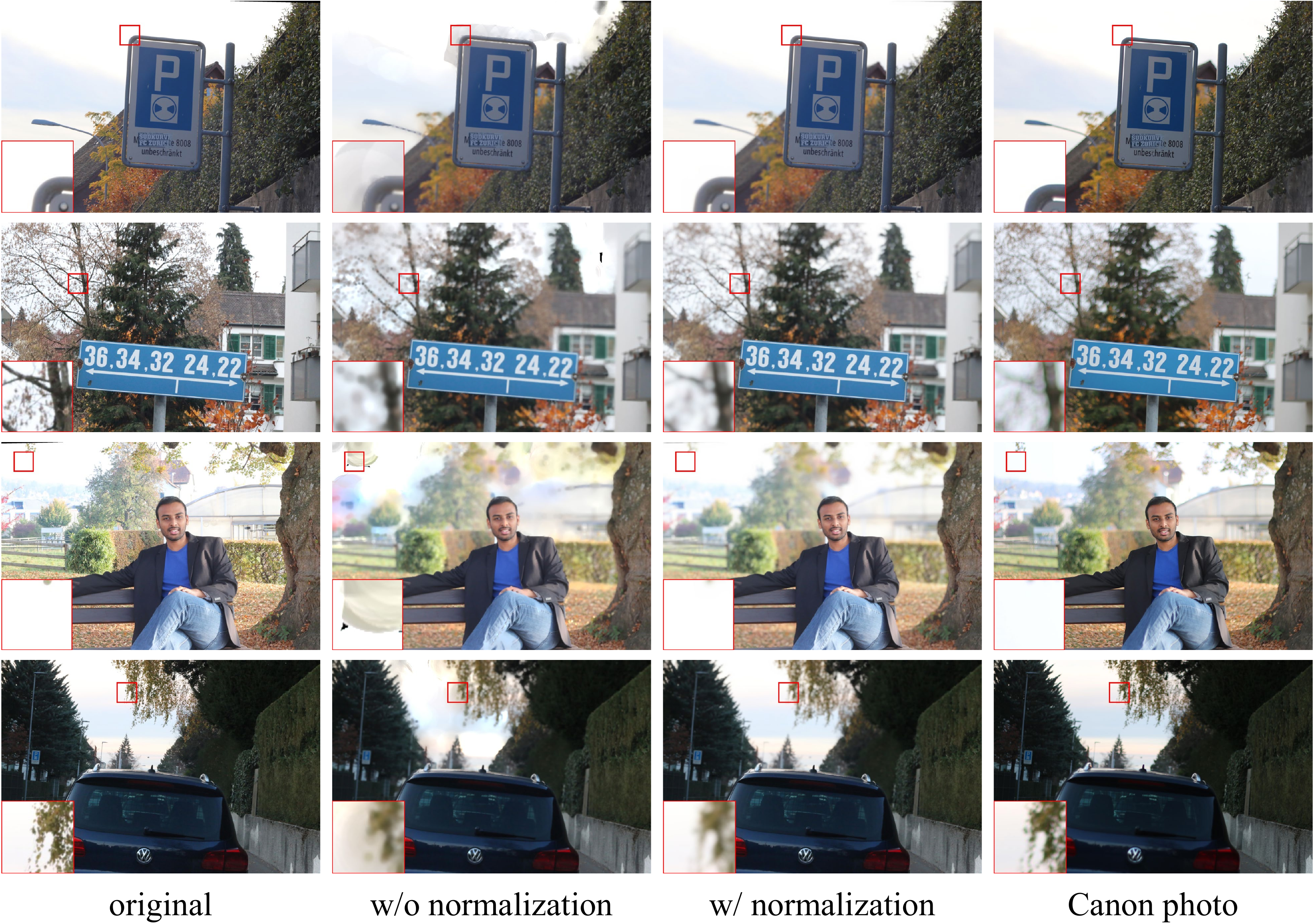}
		
		\caption{Qualitative results with or without normalization.}
		\label{fig:normal}
	\end{figure}

	Here we add the subscript to $\mathcal{B}$ to remind that the generated bokeh is at low resolution.
	$\mathcal{R}_r^{'}$ integrates the radiance map in bokeh rendering, which helps to produce the CoC. It is worth mentioning that layered rendering does not require training, and the formulation of the radiance map and the defocus map is learned.
    The rendering of the radiance map is introduced to ensure consistency in tones between $\mathcal{I}_{lr}$ and $\mathcal{B}_{lr}$. $f_l(\mathcal{R},\mat{D})$ acts as a normalizer of weighted layered rendering $f_l(\mathcal{R}\cdot \mathcal{I}_{lr},\mat{D})$, because we introduce the production of the radiance map and the original image, and this step might change the overall brightness. As shown in Fig.~\ref{fig:normal}, this normalization improves the quality of the rendered bokeh. In the second column, the rendered bokeh without normalization produces artifacts in areas which are supposed to be consistent in color. If we add normalization and render bokeh as the third column, the color inconsistency is eliminated.
    
    \squishend

	\subsection{Deep Poisson Fusion}\label{sec:blending}
	
    Since the rendered low-resolution bokeh $\mathcal{B}_{lr}$ needs to be recovered to the original size, we expect to find an appropriate upsampling approach. 
    Super-resolution or other learning-based methods~\cite{wang2018deeplens,wang2018esrgan} can be employed to increase the resolution of bokeh. However, these approaches tend to destroy the CoC effect. Naive upsampling methods such as bilinear upsampling can also fail to render realistic bokeh because they blur the whole low-resolution image including the in-focus regions. One may consider computing a binary mask from the defocus map to keep the clearness of the in-focus region, but we find this creates unpleasant artifacts around boundaries. Next we show how we deal with the boundary artifacts.
    
    We implement the predicted defocus map as guidance for the fusion mask. Therefore, an expected fusion pipeline defines a function $f_p$ that generates a soft mask $\mat{M}_s\in \mathbb{R}^{ {H}\times {W}}$, defined by
    \begin{equation}
        \begin{aligned}
            \mat{M}_{s}=f_p(\mathcal{I}_{hr},\mat{D})\,.
        \end{aligned}
    \end{equation}
    To predict $\mat{M}_s$, a loss function is used for fusing high-resolution small-aperture images and the upsampled rendered bokeh. 
    Inspired by Poisson blending loss~\cite{zhang2020deep}, the Poisson gradient loss is applied to force the gradient consistency of the in-focus region for both large-aperture images $\mathcal{B}_{hr}\in \mathbb{R}^{{H}\times {W}\times3}$ and small-aperture images $\mathcal{I}_{hr}\in \mathbb{R}^{{H}\times {W}\times3}$. The loss function takes the form
    \begin{equation}
        L_{p}(\mathcal{B}_{t},\mathcal{B}_{hr}) = \frac{1}{2HW}\sum_{m=1}^{H}\sum_{n=1}^{W}\left [ \bigtriangledown \mathcal{B}_{t} -\bigtriangledown \mathcal{B}_{hr} \right ]_{mn}^{2}\,,
    \end{equation}
    where $\mathcal{B}_{hr}=\mat{M}_s \cdot\mathcal{I}_{hr} + (1-\mat{M}_s)\cdot\mathcal{B}_{up}$ is the final fusing result. $B_{t}$ represents the target image and is defined by
    \begin{equation}
    	\mathcal{B}_t=\mathcal{I}_{hr}\cdot \mat{M}_b+\mathcal{B}_{up}\cdot  (1-\mat{M}_b)\,,
    	\label{target}
    \end{equation}
    where $\bigtriangledown$ indicates the Laplacian gradient operator. $\mathcal{B}_{up}\in \mathbb{R}^{ {H}\times {W}\times3}$ is generated from $\mathcal{B}_{lr}$ by bilinear upsampling.
    $\mat{M}_b\in \mathbb{R}^{{H}\times {W}}$ is a binary mask converted from the defocus map using a threshold $\theta$ such that, for any position $(x,y)$, $\mat{M}_b(x,y)=1$ if $\mat D(x,y)>=\theta$, and $\mat{M}_b(x,y)=0$ if $\mat D(x,y)<\theta$. 
    We integrate the defocus map in our fusion framework as Eq.~\eqref{target} to guide the deep Poisson network.

    \subsection{Loss Function}\label{subsec:Loss function}

    The training of the network is divided into two stages. The first stage jointly trains defocus hallucination and radiance virtualization, while the second stage trains deep Poisson network. Layered rendering does not require training.
    
    The loss for defocus hallucination and radiance virtualization is defined by
    \begin{equation}\label{eq:baseloss}
    \begin{aligned}
        L_{stage1} = L_{1}(\mathcal{B}_{hr},\mathcal{B}_{gt}) + \lambda\cdot L_{vgg}(\mathcal{B}_{hr},\mathcal{B}_{gt}) \\+  L_{ssim}(\mathcal{B}_{hr},\mathcal{B}_{gt}) + \delta\cdot L_{grad}(\mat{D},\mathcal{I}_{lr})
    \end{aligned}\,,
    \end{equation}
    where $L_{1}$ is the $\ell_1$ loss, $L_{vgg}$ is the perceptual loss based on the pre-trained VGG19~\cite{johnson2016perceptual}, and $L_{ssim}$ is the structural similarity (SSIM) loss~\cite{DBLP:journals/tci/ZhaoGFK17}.
    $L_{grad}$ is the pyramid gradient loss used to constrain the defocus map to be locally smooth, especially in the areas with consistent colors. $L_{grad}$ takes the form
    \begin{equation}
    \begin{aligned}
        L_{grad}(\mat{D},\mathcal{I}_{lr}) = \frac{1}{S}\sum_{i=1}^S\bigg(\left|\partial_x \mat{D}^i\right|\cdot e^{-\left\|\partial_x \mathcal{I}_{lr}^i\right\|_1} \\ 
        +\left|\partial_y \mat{D}^i\right|\cdot e^{-\left\|\partial_y \mathcal{I}_{lr}^i\right\|_1}\bigg)
    \end{aligned}\,,
    \end{equation}
    where $S$ denotes the the number of scales.
	
	For training deep Poisson network, 
	we further add the poisson gradient loss to Eq.~\eqref{eq:baseloss} as
	\begin{equation}
    \begin{aligned}\label{eq:totalloss}
        L_{stage2} = L_{stage1} + \zeta \cdot L_{p}(\mathcal{B}_{t},\mathcal{B}_{hr})\,.
    \end{aligned}
    \end{equation}
    The Poisson gradient loss adds gradient constraints to in-focus areas, as in Fig.~\ref{fig:pipeline}.

    \subsection{Implementation Details}\label{Implementation Detail}
    We implement our method based on \texttt{PyTorch}. The hardware platform includes a single Nvidia GTX $1080$ GPU with $256$ GB RAM and Intel Xeon processor. To accelerate the inference speed, we use ResNeXt50~\cite{xie2017aggregated} pre-trained on ImageNet~\cite{deng2009imagenet} as the backbone in defocus hallucination. In Eq.~\eqref{eq:hdr}, empirically $\alpha=3$ and $\beta=5$, because we want to widen the difference between bright pixels that are considered in HDR and other pixels. The threshold $\theta$ used to generate $\mat{M}_b$ is set to $0.25$, because we find that the threshold does not have a decisive effect on the generated bokeh. In Eq.~\eqref{eq:smooth}, $\gamma=100$, because a smaller value tends to highlight the gap between each layer and make the result more fragmented, and a larger value would fade the CoC. In Eq.~\eqref{eq:softkernel}, $\sigma=0.25$ and $\varphi=0.5$, because under this setting, the blur kernel resembles a smooth disk. In Eq.~\eqref{eq:baseloss}, $\lambda$ is set to $0.1$ because we want to keep $L_{1}(\cdot,\cdot)$, $L_{vgg}(\cdot,\cdot)$, and $L_{ssim}(\cdot,\cdot)$ on the same order of magnitude at the beginning of training. $\delta$ is set to $0.1$, so it is enough to restrain the gradient of defocus map while maintaining the quality of generated bokeh. The number of scale $S$ in the pyramid gradient loss is set to $4$. In Eq.~\eqref{eq:totalloss}, $\zeta$ is set to $10$, so it is enough to ensure the sharpness of in-focus regions.

	\section{Results and Discussions}\label{sec:results}

	In this section, we first introduce the experimental setting. Then we report the quantitative and qualitative performance of \ourmethod on a large-scale bokeh dataset EBB!. Finally, we conduct an ablation study to analyze the influence of different factors on bokeh.

	\subsection{Dataset and Experimental Setup}
    {\it Everything is Better with Bokeh!} (EBB!)~\cite{ignatov2020rendering} is a large-scale dataset consisting of $4694$ aligned wide/shallow depth-of-field image pairs captured using the Canon 7D DSLR with 50mm f/1.8 lenses. EBB! is initially proposed for the AIM 2020 Rendering Realistic Bokeh Challenge~\cite{ignatov2020aim}. The purpose of this challenge is to achieve the shallow depth-of-field with the best perceptual quality similar to the ground truth as measured by the Mean Opinion Score (MOS). MOS is defined as a numeric value ranging from 1 to 5 (5 - comparable perceptual quality, 4 - slightly worse, 3 - notably worse, 2 - poor perceptual quality, 1 - completely corrupted image)~\cite{ignatov2020aim}.
    For evaluation, we divide the dataset into a training set with $4224$ pairs and a validation set with $470$ pairs, which is termed as Val470. The training process includes two stages. 
    At the first stage, we jointly train the defocus hallucination module and the radiance virtualization module on the $512\times512$ image resolution; 
    at the second stage, we train the deep Poisson fusion alone on the original $1024\times1024$ resolution.
    
    In experiments, we use the peak signal to noise ratio (PSNR) and structural index similarity (SSIM) for evaluation. PSNR and SSIM are two widely used metrics in image quality assessment. In addition, MOS is the quantitative criterion in AIM 2020 Rendering Realistic Bokeh Challenge.
	
	\subsection{Comparison With State of the Art}
	
    \subsubsection{Quantitative Results}
    \begin{figure*}
        \centering
        \includegraphics[width=1.0\linewidth]{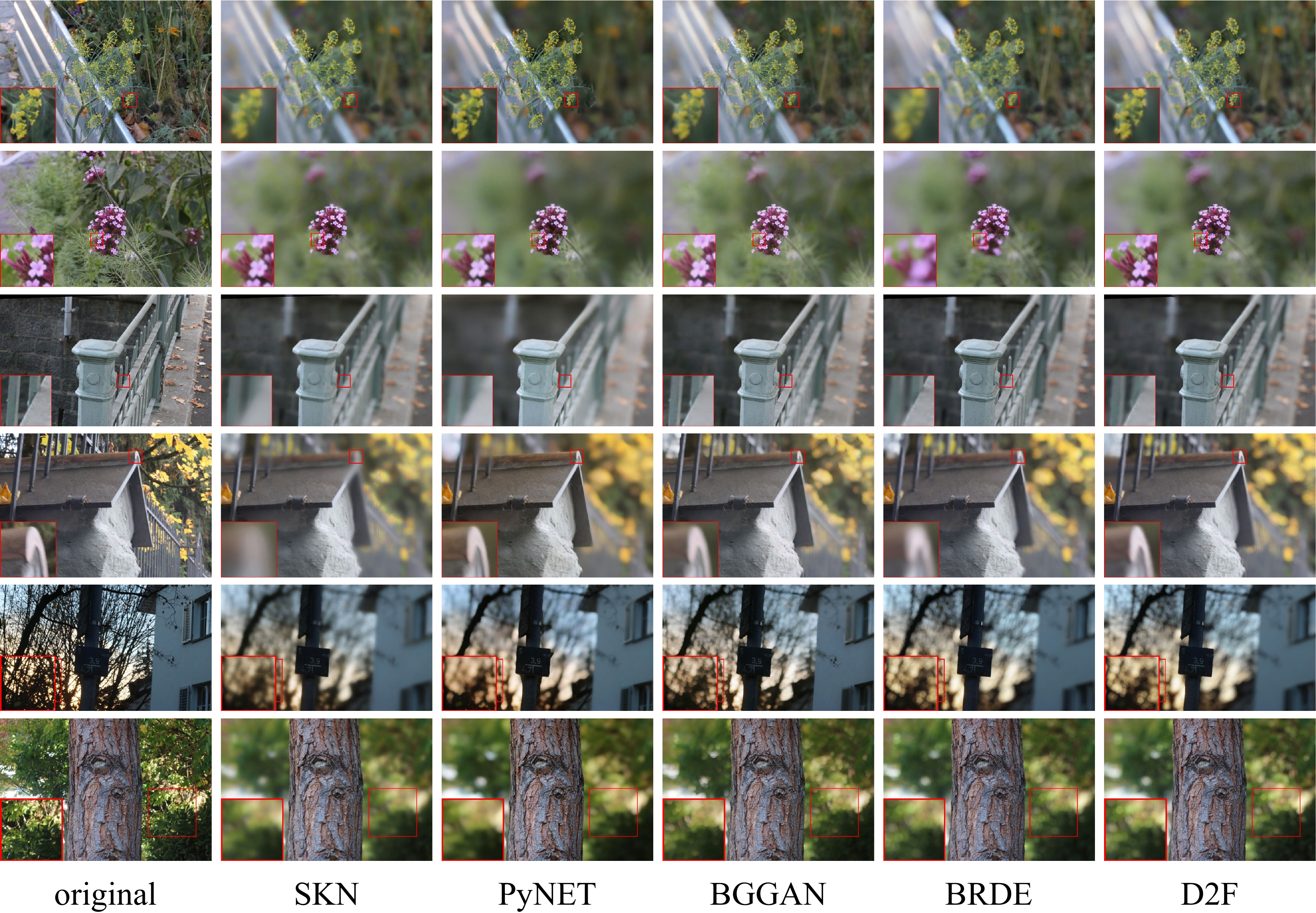}

        \caption{ \textbf{Visual results obtained with four different methods.} From left to right: the original narrow aperture image, SKN(Selective Kernel Networks for Bokeh Effect Simulation) which ranked fourth from Track 1 of AIM 2019 Challenge on Bokeh Effect Synthesis~\cite{ignatov2019aim}, PyNET~\cite{ignatov2020rendering}, BGGAN~\cite{qian2020bggan}, our previous naive solution~\cite{luo2020bokeh}
        , and our current gradient-based fusion solution. In the first four rows, \ourmethod achieves better results than our previous solution~\cite{luo2020bokeh}, and \ourmethod is competitive among other methods. In addition, in the last two rows, \ourmethod obtains prominent CoC compared with other methods.}
    \label{fig:comparetest}
    \vspace{-10pt}
    \end{figure*}
    \ourmethod is initially proposed to participate in the AIM 2020 Rendering Realistic Bokeh Challenge~\cite{ignatov2020aim} and our solution is the runner-up of all methods. In the competition, we apply a naive alpha blending approach for upsamping~\cite{luo2020bokeh}.
    
    Table~\ref{tab:aimquantitative} shows the performance of our model. We compare results from AIM 2019 Rendering Realistic Bokeh Challenge~\cite{ignatov2019aim} as well as AIM 2020 Rendering Realistic Bokeh Challenge. 
    It is worth mentioning that we send our new result to the organizers and get PSNR and SSIM results, however, the corresponding MOS result is unlikely to obtain because it is from a user study that is already finished. 
    As shown in Table~\ref{tab:aimquantitative}, \ourmethod with deep Poisson fusion achieves better results than our previous method~\cite{luo2020bokeh} in terms of PSNR and SSIM, and our previous method
    ranks second in AIM 2020 Rendering Realistic Bokeh Challenge.
    Although we are not able to compare \ourmethod with methods from AIM 2020 Rendering Realistic Bokeh Challenge on MOS, we provide qualitative results to show that \ourmethod is superior to our previous work.

    \begin{table}[!t] \small
        \centering
        \captionsetup{singlelinecheck=false,labelsep=newline,justification=centering}
        \caption{\scshape Quantitative Results From the AIM 2020 Rendering Realistic Bokeh Challenge. Results Are Sorted Based on the MOS scores. Our Current Method \ourmethod and Our Previous Method Are in the Lavender Rows. The Best Performance Is in Boldface}
        \label{tab:aimquantitative}
        \begin{tabular}{cccc}
            \toprule
            Team&MOS&PSNR & SSIM\\
            \hline
            BGGAN~\cite{qian2020bggan}&\textbf{4.2}&23.58&0.8770\\
            PyNET~\cite{ignatov2020rendering}&--&23.28&0.8780\\
            
            SKN~\cite{ignatov2019aim}&--&23.18&0.8851\\
            
            Dutta \textit{et al.}~\cite{dutta2021depth}&--&22.14&0.8633\\
            \rowcolor{Lavender} \ourmethod& -- &23.11&0.8862\\
            \rowcolor{Lavender} BRDE~\cite{luo2020bokeh}&4.0&22.94 & 0.8842 \\
            CET\_SP~\cite{ignatov2020aim}&3.3&21.91&0.8201\\
            CET\_CVLab~\cite{ignatov2020aim}&3.2&23.05&0.8591\\
            TeamHorizon~\cite{ignatov2020aim}&3.2&23.27&0.8818\\
            IPCV\_IITM~\cite{ignatov2020aim}&2.5&\textbf{23.77}&\textbf{0.8866}\\
            CET21\_CV~\cite{ignatov2020aim}&1.3&22.80&0.8628\\
            CET\_ECE~\cite{ignatov2020aim}&1.2&22.85&0.8629\\
            \bottomrule
        \end{tabular}
    \end{table}

    \subsubsection{Qualitative Results}
    
    In this section, we compare \ourmethod to the state-of-the-art solutions~\cite{ignatov2019aim,ignatov2020rendering,qian2020bggan} that were trained and tuned specifically for bokeh rendering. The visual results of all methods are presented in Fig.~\ref{fig:comparetest}. 
    As shown in Fig.~\ref{fig:comparetest}, \ourmethod shows competitive results in producing CoC. In addition, \ourmethod achieves better quality compared with the previous naive blending method, because the manual threshold of naive blending sometimes does not fit the minimum value of defocus map, in which case the naive blending results have a small clear area, even none. Implementing Poisson blending solves this issue.
    \begin{table}[!t] \small
        \centering
        \captionsetup{singlelinecheck=false,labelsep=newline,justification=centering}
        \caption{\scshape Runtime of Different methods on GPU}
        \label{tab:runtime}
        {\begin{tabular}{cccc}\toprule
             \small
             &SKN~\cite{ignatov2019aim}&\ourmethod&PyNet~\cite{ignatov2020rendering}\\\hline
            
             Time(s)& 0.87 &0.58 &0.24\\
             
             \bottomrule
        \end{tabular}}
    \end{table}
    
    \subsubsection{Runtime}
    In this section we compute time in ms of different methods on GPU. The result is shown in Table~\ref{tab:runtime}.
    \begin{figure*}[!htb]
        \centering
        
        \includegraphics[width=1.0\linewidth]{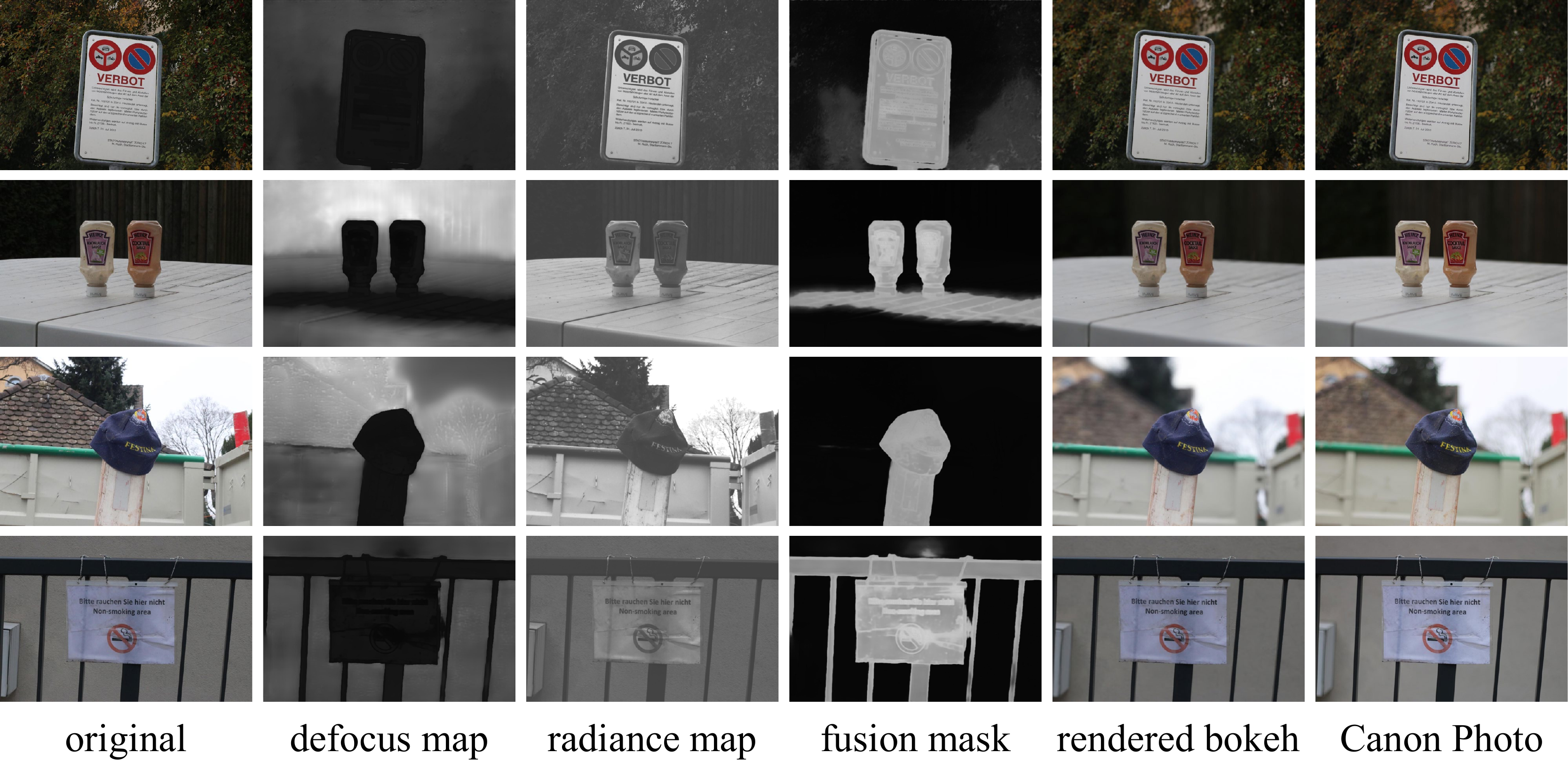}
        
        \caption{ \textbf{Visual results obtained with \ourmethod}. Best zoomed-in on screen.}
        \label{fig:intermediate}
        \vspace{-10pt}
        \end{figure*}
    
    \subsection{Ablation Study}
    As mentioned above, our model consists of different components and is trained in a multi-module manner with the combination of different losses and multiple blur kernels. Therefore, we conduct an ablation study to compare different settings and demonstrate the effectiveness and superiority of \ourmethod. 
    Fig.~\ref{fig:intermediate} presents intermediate outputs which consist of defocus maps, weight maps, fusion masks, and the generated bokeh.

    \begin{table}[!t]
        \centering
        \captionsetup{singlelinecheck=false,labelsep=newline,justification=centering}
        \caption{\scshape Quantitative Results of Defocus Hallucination and Radiance Virtualization Training With Different Losses on Val470}
        \label{tab:loss}
            \begin{tabular}{rcc}\toprule

            \small
            Loss & PSNR & SSIM\\\hline
            $L_{l1}$ & 23.6241 & 0.8730  \\
            $L_{l1}\!+\!L_{grad}$ &23.6202& 0.8745  \\
            $L_{l1}\!+\!L_{ssim}\!+\!L_{grad}$ &23.6780&0.8794   \\
            $L_{total}$ &\textbf{23.7091} & \textbf{0.8795}\\
            \bottomrule
            \end{tabular}
    \end{table}
    
    \begin{table}[!t]
        \centering
        \captionsetup{singlelinecheck=false,labelsep=newline,justification=centering}
        \caption{\scshape Quantitative Results of Fusion Network Training With Different Losses on Val470}
        \label{tab:blendloss}
            \begin{tabular}{rcccc}\toprule
            \small
            Loss & PSNR & SSIM\\\hline
            $L_{l1}\!+\!L_{Poisson}$ & 23.6605 & 0.8776  \\
            $L_{l1}\!+\!L_{ssim}\!+\!L_{Poisson}$ &23.7123& 0.8817  \\
            $L_{l1}\!+\!L_{ssim}\!+\!L_{vgg}\!+\!L_{Poisson}\!$ &23.7298&0.8817   \\
            $L_{total}$ &\textbf{23.7342} & \textbf{0.8818}\\
            \bottomrule
            \end{tabular}
    \end{table}

    \subsubsection{Combination of Losses} 
\begin{figure*}[!htb]
    \centering
    \includegraphics[width=1.0\linewidth]{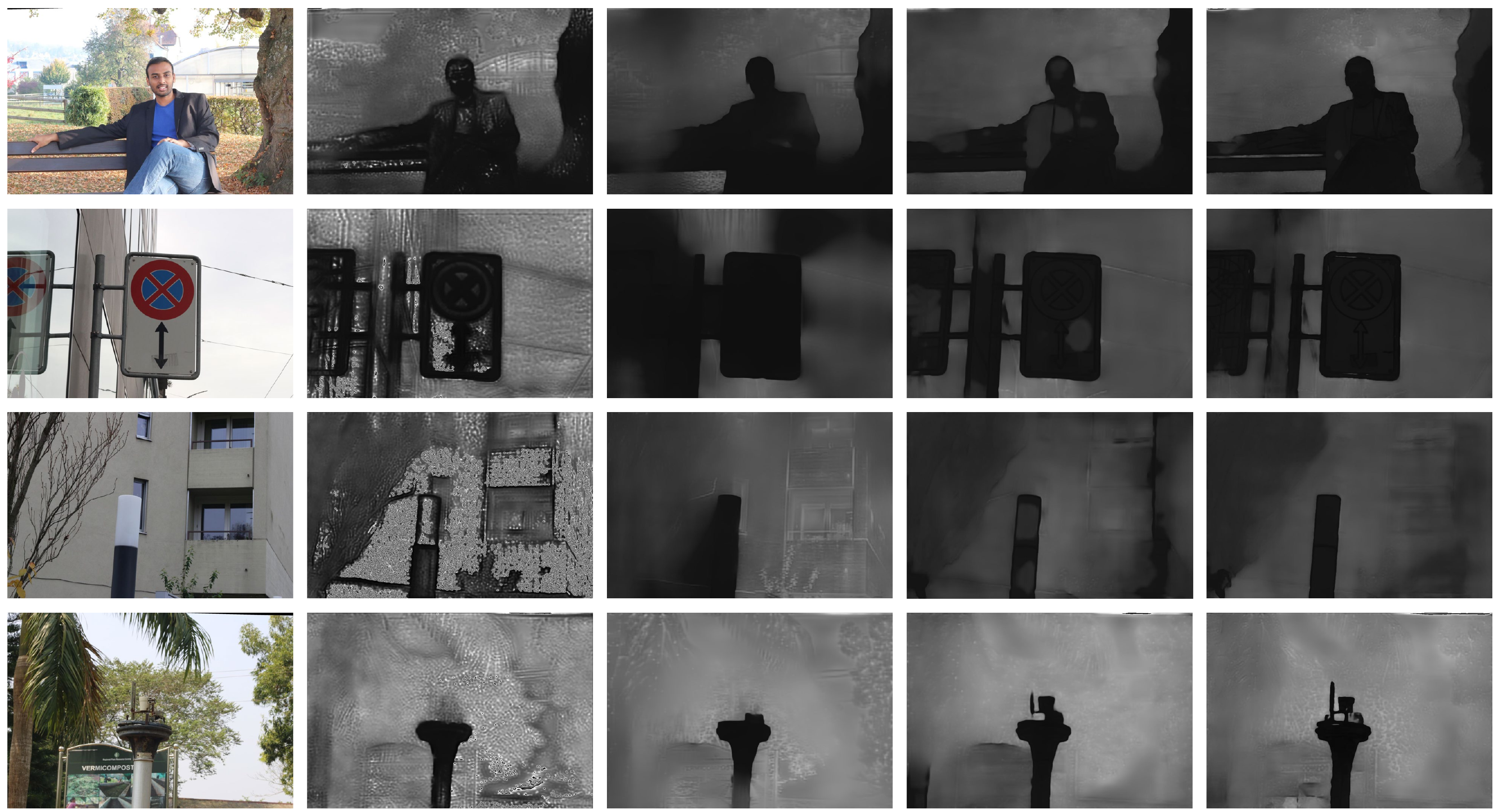}
    \caption{
    \textbf{Visualization of defocus maps.} The first row is all-in-focus image. The second row to the last row present the predicted defocus maps training with $L_{l1}$, $L_{l1}\!+\!L_{grad}$, $L_{l1}\!+\!L_{ssim}\!+\!L_{grad}$ and $L_{total}$, respectively.
    }
    \label{fig:grad}
    \end{figure*}

As shown in Table~\ref{tab:loss} and Table~\ref{tab:blendloss}, we obtain the best results when all of the loss functions are used. Besides, we visualize some examples of the predicted defocus maps with different combinations of losses in Fig.~\ref{fig:grad} to verify the effect of each loss. One can see that the predicted defocus maps become more delicate on in-focus regions, smoother on out-of-focus regions, and more precise in terms of blurring amount by adding loss function gradually. The third column indicates that the gradient loss helps to generate consistent defocus maps with sharp discontinuities around in-focus objects such as the human (row 1), the sign (row 2) and the pole (row 3,4). If we add SSIM loss as the fourth column, the quality of defocus map is improved in background regions, and the shape of objects is better retained, such as the pole and the man. The fifth column shows that the perceptual loss helps to improve the overall quality of the defocus map, as the man, the sign and the pole are all intact, and the background is flattened.
    Particularly, we discovered that pyramid gradient loss in Fig.~\ref{fig:grad} is helpful for the training process to converge. Not only does pyramid gradient loss make the predicted defocus map more accurate, but also it improves the performance of generated bokeh. 

    \subsubsection{Settings of Blur Kernels} 
\begin{figure}[!t]
        \centering
    
        \subfloat[a]{\includegraphics[width=2.25in]{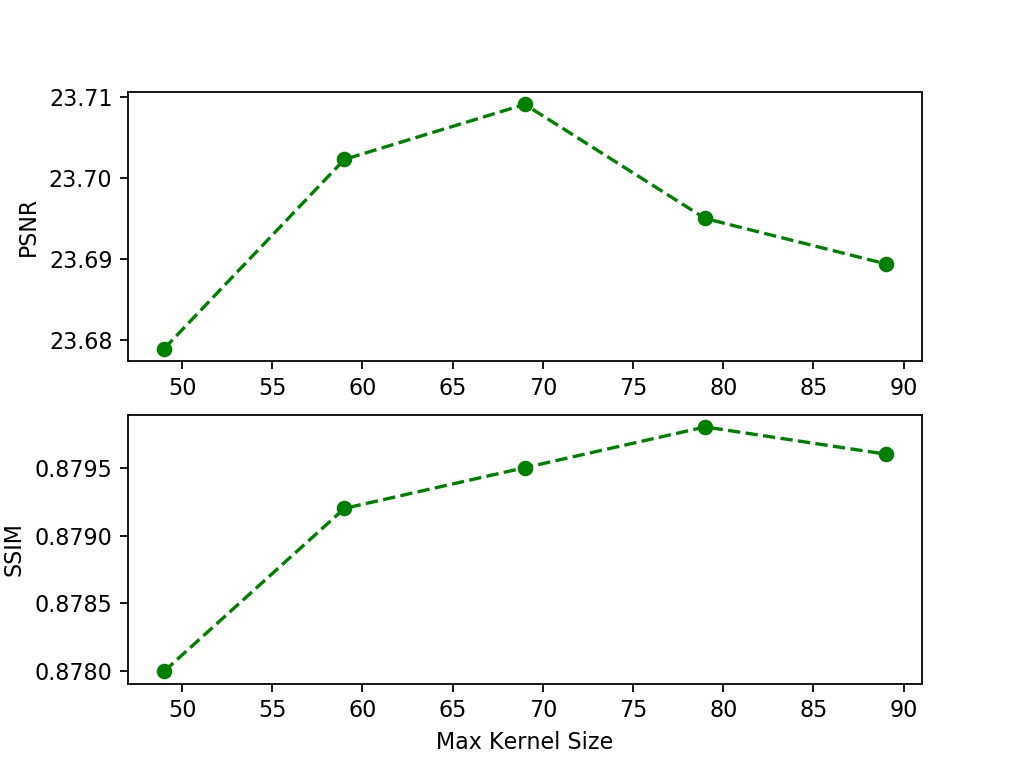}}
        \subfloat[b]{\includegraphics[width=2.25in]{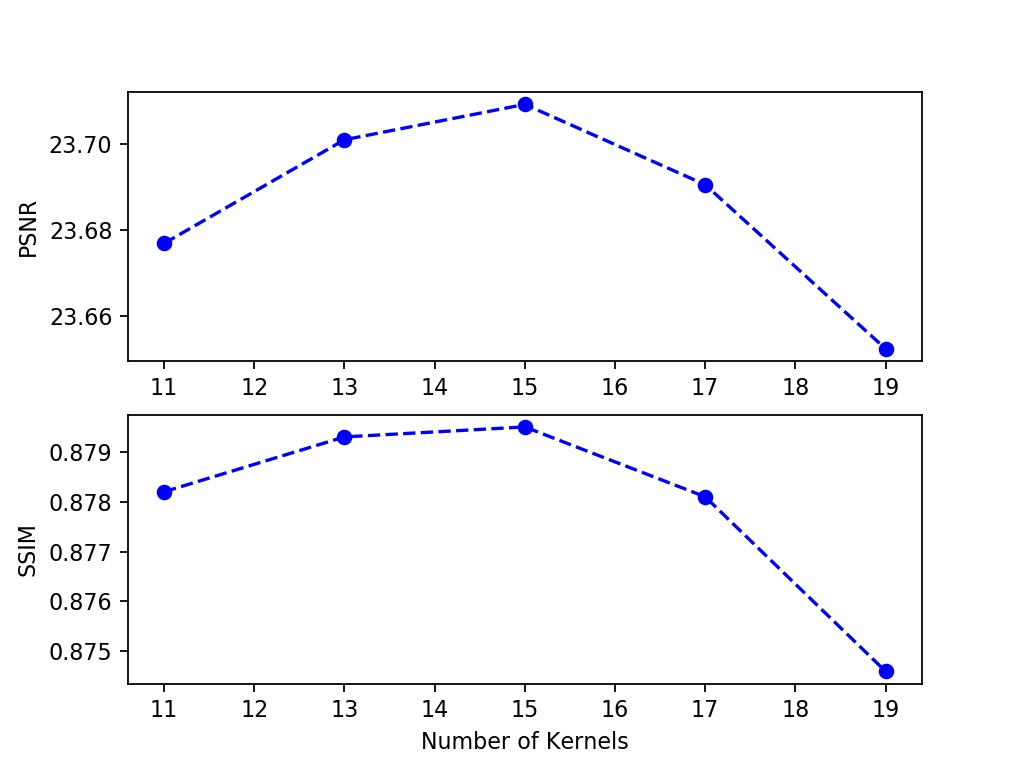}}

        \caption{
        PSNR and SSIM results of different maximum kernel sizes and different number of kernels. }
        \label{fig:kernelablation}
        \end{figure}

The number and size of the blur kernels are largely determined by experience. On the one hand, too many kernels will slow down the running speed while too few kernels will cause a certain limitation. On the other hand, the maximum kernel size is supposed to be consistent with the maximum blur amount in a real scene. However, we observe that the defocus map corresponding to the large-scale blur is hard to learn as the blur amount varies greatly among the images and most of them have a small blur amount. In consequence, we choose to have more blur kernels with a small blur radius than those with a large blur radius. In addition, We conduct some experiments on the settings of blur kernels. We compare the different numbers and maximum sizes of blur kernels in Fig.~\ref{fig:kernelablation}. We chose 11, 13, 15, 17, and 19 kernels for ablation studies. 

    We adopt the 15-kernel solution in Eq.~\eqref{eq:15kernel} according to the experiments. 
    In addition, we choose the maximum kernel sizes to be 49, 59, 69, 79, and 89. 
    We adopt the maximum kernel size to be 69 as Eq.~\eqref{eq:15kernel} according to the experiment to achieve better PSNR and SSIM and occupy less memory. To generate a smooth bokeh result, the kernel size should be set continuously. We adopt two sampling strategies, i.e., growing sampling and uniform sampling. The growing sampling is what we adopt in Section~\ref{sec:bokeh_rendering}. For uniform sampling, the interval of kernel sizes is set to 4 from beginning to end. The comparison between the two strategies is shown in Table~\ref{tab:kernel_sample}. The above analysis can help us set the pre-defined blur kernels better.
    
    \begin{table}[!t] \small
        \centering
        \captionsetup{singlelinecheck=false,labelsep=newline,justification=centering}
        \caption{\scshape Quantitative Results of Different Sampling Strategies on Val470}
        \label{tab:kernel_sample}
        \begin{tabular}{ccc}\toprule
            \small
            ~& Growing & Uniform \\\hline
    
            PSNR & \textbf{23.7091} & 23.5260 \\
            SSIM & \textbf{0.8795} & 0.8628 \\
            \bottomrule
        \end{tabular}
    \end{table}

    \begin{figure}[!htb]
        \centering
        \includegraphics[width=1.0\linewidth]{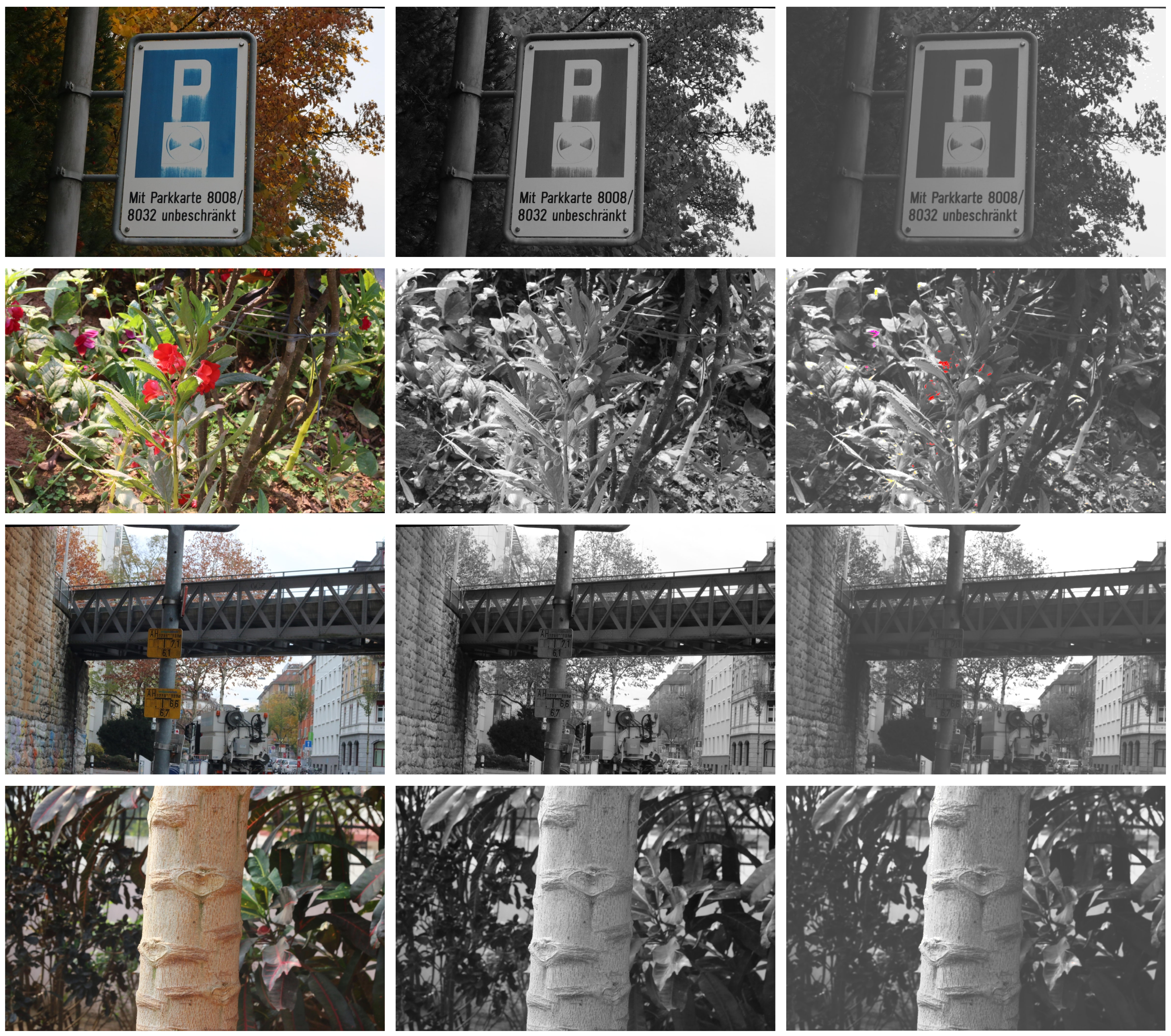}
        \caption{
        Qualitative results of gray images and radiance maps.}
        \label{fig:weight}
        \end{figure}

	\begin{figure}[!htb]
        \centering
        \includegraphics[width=1.0\linewidth]{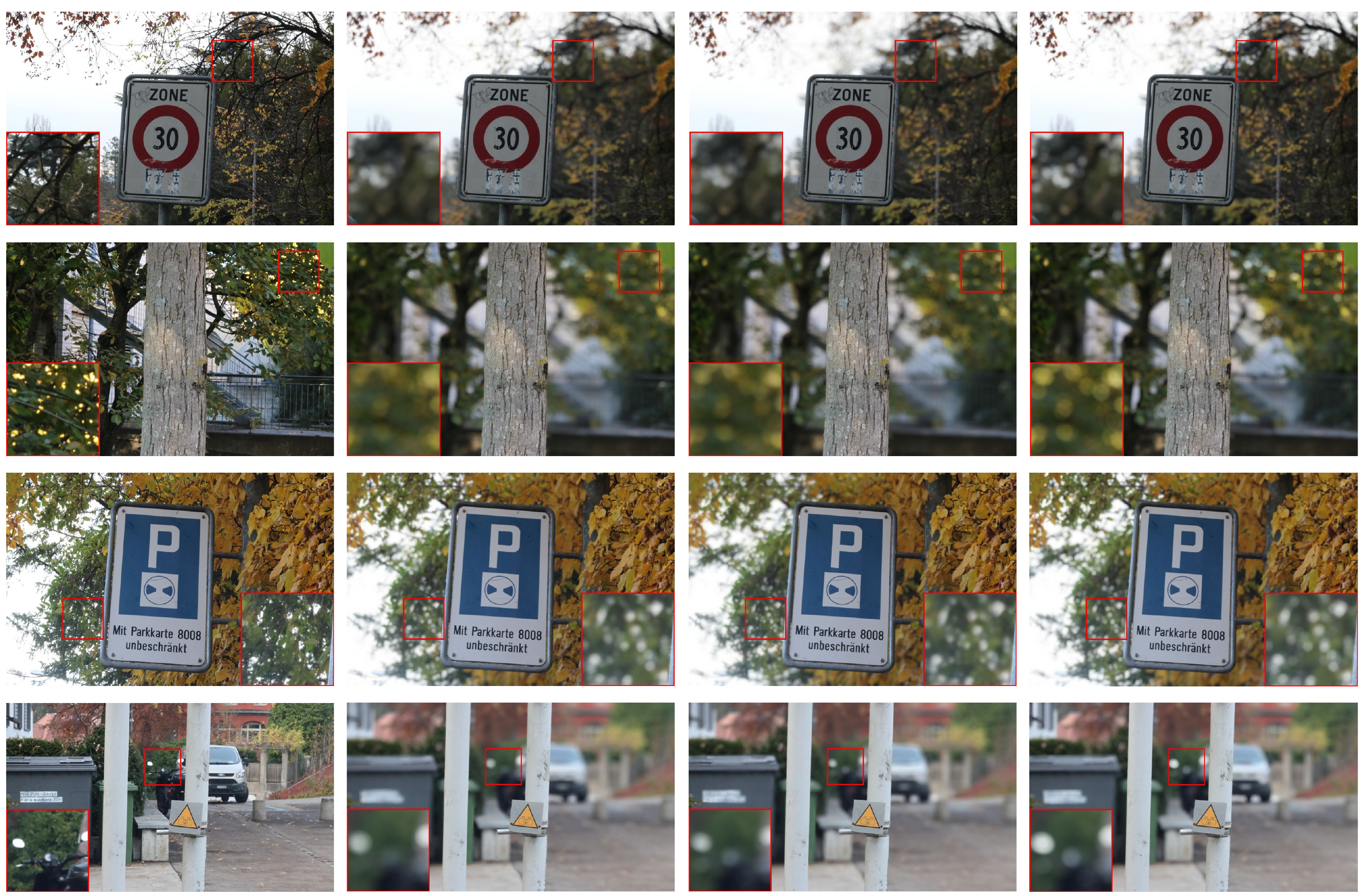}
        \caption{
        Qualitative results of with or without radiance maps, or applying gray images.}
        \label{fig:radiance}
        \end{figure}
    \begin{figure*}[!htb]
        \centering
        \includegraphics[width=1.0\linewidth]{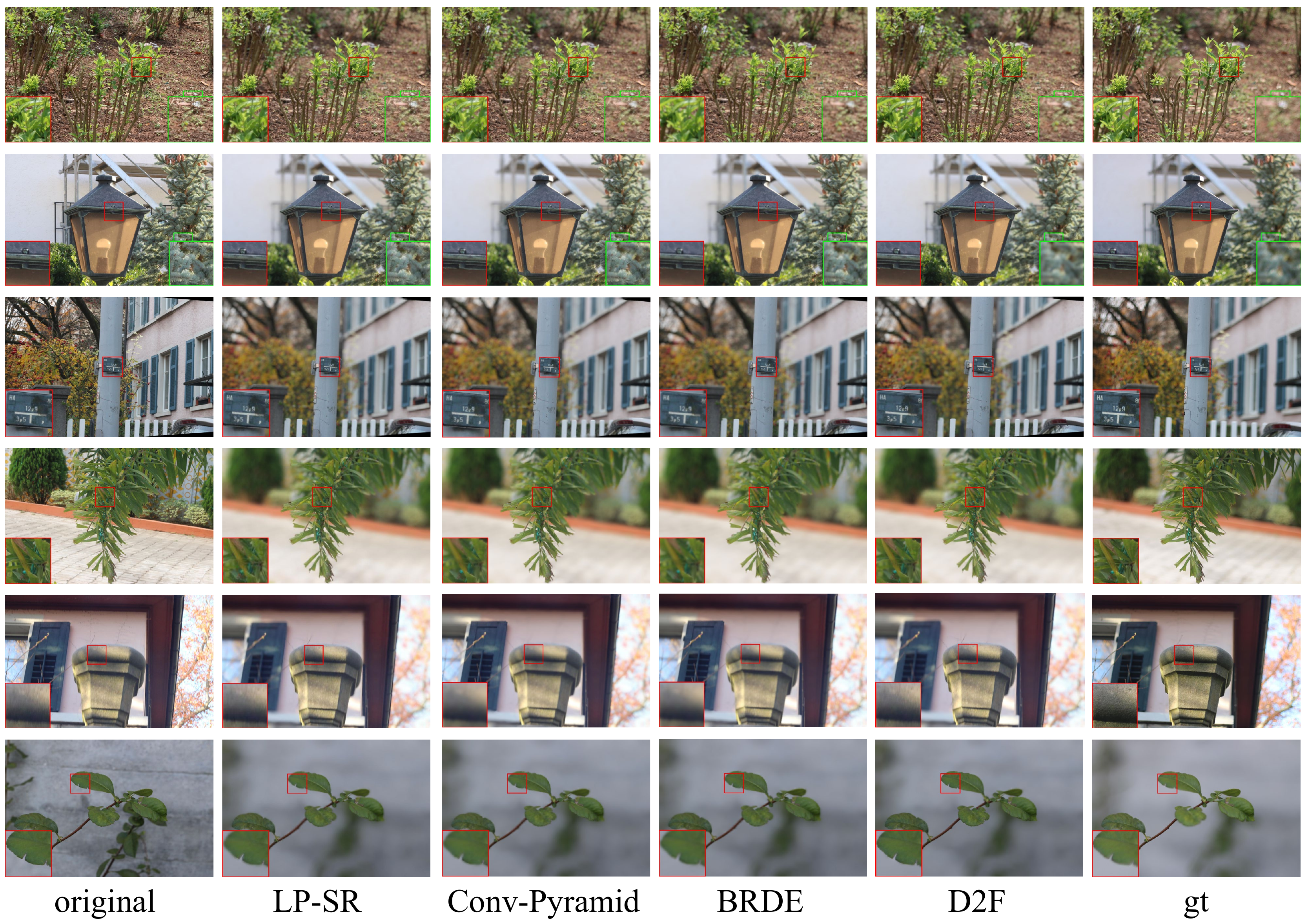}
        \caption{
        \textbf{Qualitative results of different fusion methods.} From left to right: the all-in-focus input, LP-SR~\cite{liu2015general}, Conv-Pyramid~\cite{farbman2011convolution}, our previous solution~\cite{luo2020bokeh}, \ourmethod and the ground truth. Green square and red square in each image represent respectively inconsistent blurring and blurred in-focus regions. Our method is the best solution compared to other approaches regarding the two standards. }
        \label{fig:blend}
        \end{figure*}
    
    \subsubsection{Radiance Module} As shown in Table~\ref{tab:radiance_module}, the model with radiance virtualization achieves better results with the model without radiance module quantitatively. The obtained radiance map can demonstrate scene radiance, which is more suitable for bokeh rendering than image intensity. To further prove this, we show some examples in Fig.~\ref{fig:weight} and Fig.~\ref{fig:radiance}, where we present differences with or without radiance relationship as well as the gap between radiance maps and gray images generated from RGB images.
    \begin{table}[!t] \small
        \centering
        \captionsetup{singlelinecheck=false,labelsep=newline,justification=centering}
        \caption{\scshape Quantitative Results With or Without Radiance Virtualization, or Using Gray Images as Radiance Guidance on Val470}
        \label{tab:radiance_module}
        {\begin{tabular}{cccc}\toprule
             \small
             &w/ Radiance & w/o Radiance&Gray image\\\hline
            
             PSNR&\textbf{23.7091}&23.6780&23.6527\\
             SSIM&\textbf{0.8795}  &0.8794&0.8774\\
             \bottomrule
        \end{tabular}}
    \end{table}
    
    In Fig.~\ref{fig:weight} we can distinguish between gray images and radiance maps which demonstrate scene radiance, and in Fig.~\ref{fig:radiance} bokeh results from a radiance map and a gray image are presented, we can conclude that a radiance map help produce CoC in the blurred out-of-focus regions with its revelation of radiance relationships between pixels, while bokeh obtained from a gray image or only an original RGB input fails to achieve distinctive CoC effects. 
    
    \subsubsection{Fusion Methods} 
    Our goal is to maintain the clearness of in-focus regions, and we adopt various fusion approaches to achieve our expectations. Our criteria can be divided into two parts: visual clearness of in-focus objects and overall PSNR and SSIM. In other words, we aim to reach better results quantitatively and qualitatively. In Table~\ref{tab:comparison for blending}, we present six schemes.
    Naive blending represents the method applying in~\cite{luo2020bokeh}. 
    For Laplacian Pyramid blending, we generate the mask in the same manner as naive blending. 
    For naive Poisson blending, we transform defocus map into a binary mask, which is seen as input together with the original narrow-aperture image to produce blending results. As shown in implementation details, the threshold for transformation is set to 0.25, where any pixel with a defocus value larger than 0.25 is set to 0. In addition, we evaluate two recent fusion methods. One of the methods is based on multi-scale transform and sparse representation~\cite{liu2015general}, and the other one is based on convolution pyramid blending(Conv-Pyramid)~\cite{farbman2011convolution}. One can conclude from the quantitative results that our current fusion method which employs Poisson gradient constraint and deep neural network achieves the best results.

    In Fig.~\ref{fig:blend}, we compare different fusion schemes. We discovered that bilinear upsampling is invalid for maintaining the sharpness of in-focus regions. Naive Poisson blending can produce rather clear in-focus regions. However, as we provide a binary mask set empirically to the blending process, the whole image displays inconsistent blurring regions and performs poorly around edges. Laplacian Pyramid blending performs better than naive Poisson blending on defocused regions. However, it also fails to keep sharp edges. Our network is capable of generating the most pleasing blending results among these methods.

    \begin{table}[!t]
        \centering
        \captionsetup{singlelinecheck=false,labelsep=newline,justification=centering}
        \caption{\scshape Quantitative Results of Different Fusion Methods on Val470}
        \label{tab:comparison for blending}
        {\begin{tabular}{ccc}\toprule
            \small
            Method & PSNR & SSIM \\\hline
            naive Poisson blending&23.3353&0.8708\\
            Laplacian Pyramid&23.6455 &0.8698\\              
            LP-SR~\cite{liu2015general}&23.6805&0.8713\\
            BRDE~\cite{luo2020bokeh}& 23.7091&0.8795\\
            Conv-Pyramid~\cite{farbman2011convolution}&23.7247&0.8812\\
            
            \ourmethod&\textbf{23.7342}&\textbf{0.8818}\\
            \bottomrule
        \end{tabular}}
    \end{table}

    \subsection{Training Strategy}\label{subsec:Training Strategy}

    We train our defocus hallucination module in a concise manner that the defocus hallucination network produces a single-channel defocus map on a resolution of 512$\times$512. Our deep Poisson network is trained on a resolution of 1024$\times$1024.
    \begin{figure}[!t]
        \centering
    
        \subfloat[a]{\includegraphics[width=2in]{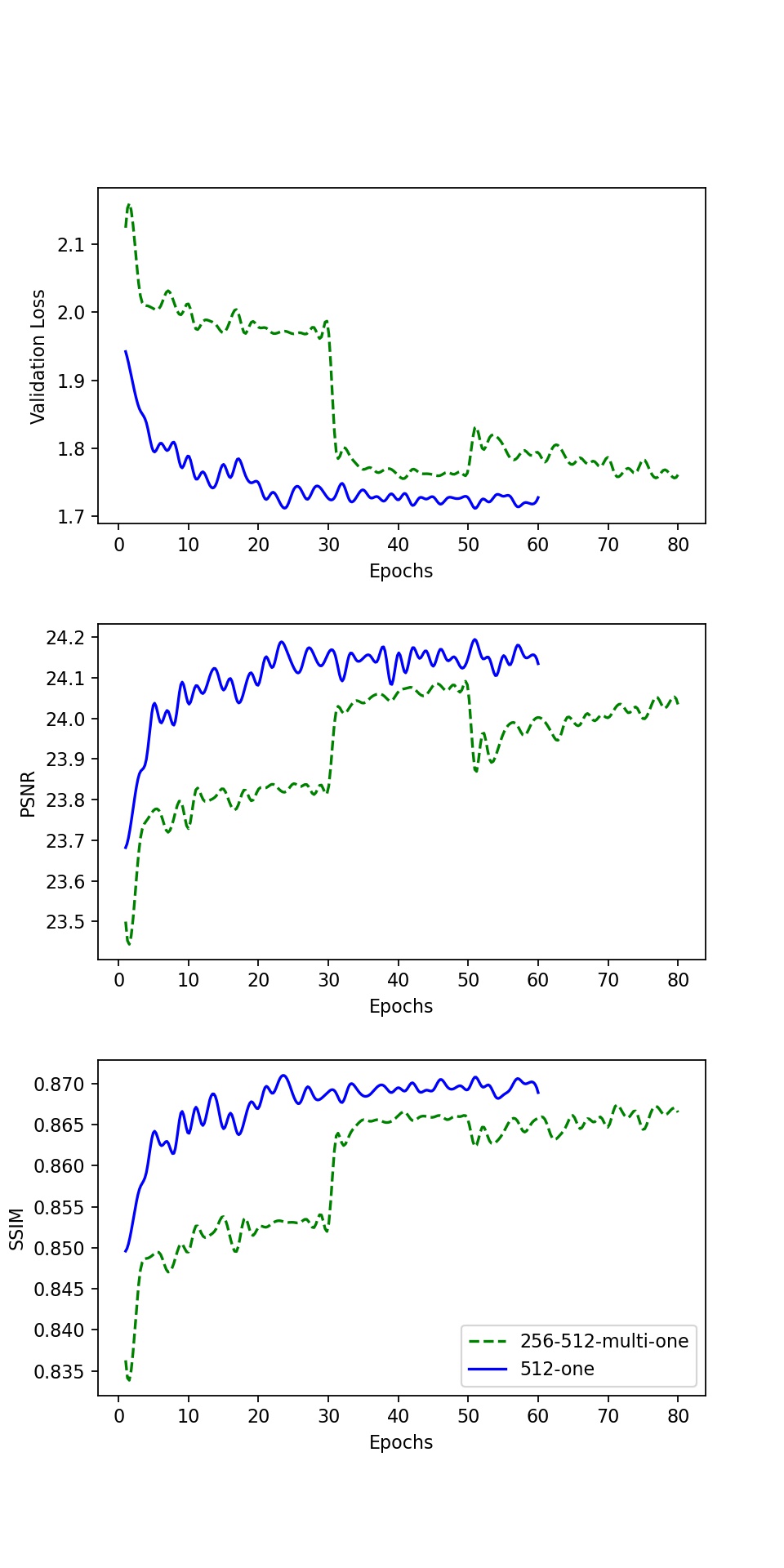}}
        \subfloat[b]{\includegraphics[width=2in]{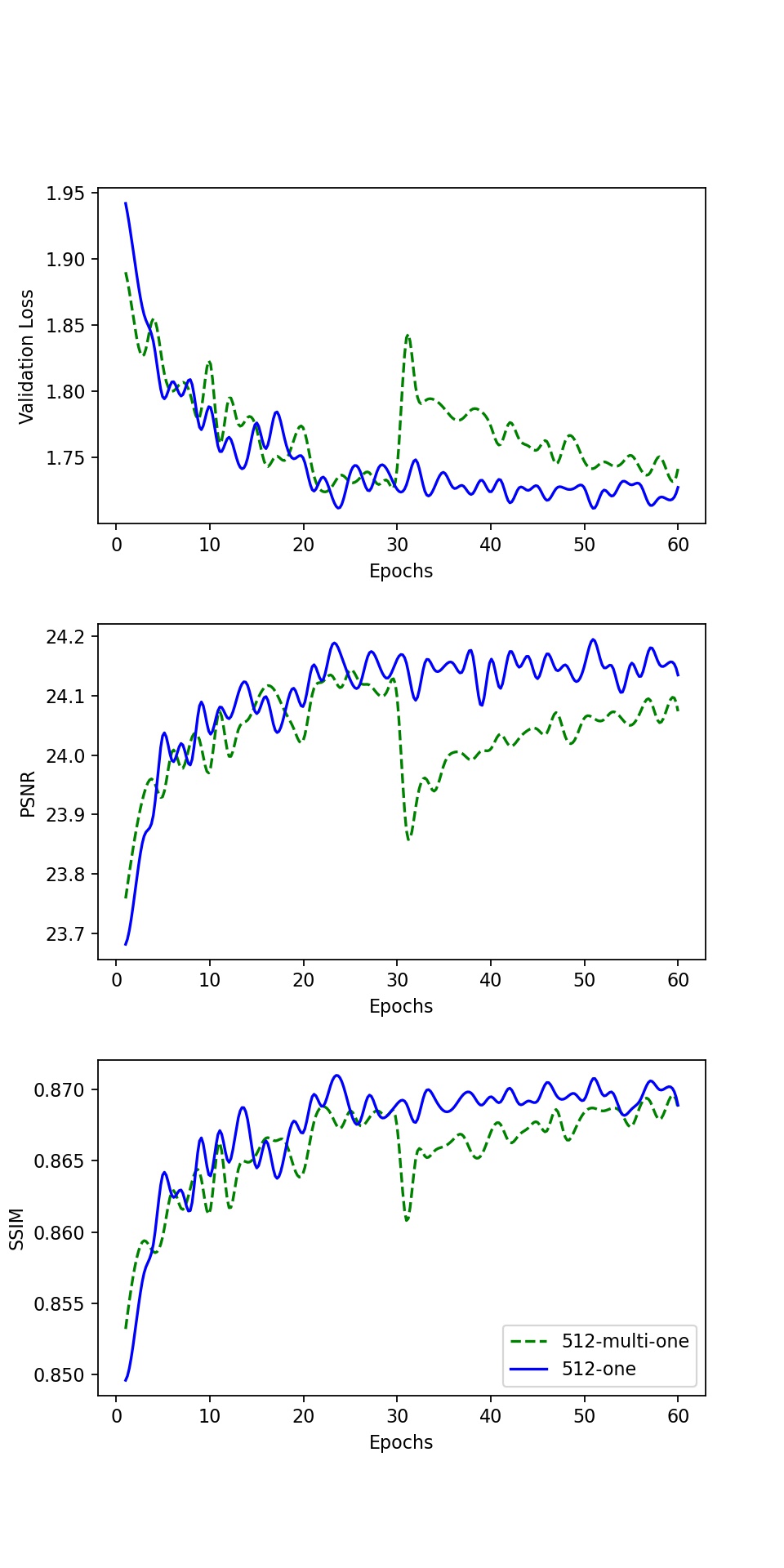}}
        
        \caption{
        Validation loss and PSNR results of different training strategy. The blue line is our current strategy.}
        \label{fig:stra}
        \end{figure}
    
    We compare our current defocus hallucination training scheme with two different strategies: i) our previous method~\cite{luo2020bokeh}, where the network is trained to produce a multi-channel defocus map on 256$\times$256 for 30 epochs, finetuned on 512$\times$512 for 20 epochs, and is trained to generate a single-channel defocus map on 512$\times$512 with the parameters of radiance virtualization module and defocus hallucination module except for the last two convolutional layers fixed for 30 epochs, and ii) A two-stage training where the network first outputs a multi-channel defocus map on 512$\times$512 to converge faster, and switches to produce the single-channel result. The multi-channel defocus map can be written as:
    \begin{equation}
        \begin{aligned}
            & \mat D_{m} = f_{m}(\mathcal I_{lr})
        \end{aligned}
    \end{equation}
    $\mat D_{m}$ is the multi-channel defocus map which is normalized by softmax function, so it can also be considered as a probabilistic map. The channels of $\mat D_{m}$ is set to 6, specifically. In order to generate single-channel results from multi-channel results, we replace the last convolutional layer and softmax layer with one convolutional layer that has a one-channel output.

    \begin{figure}[!tbp]
        \centering
        \includegraphics[width=1.0\linewidth]{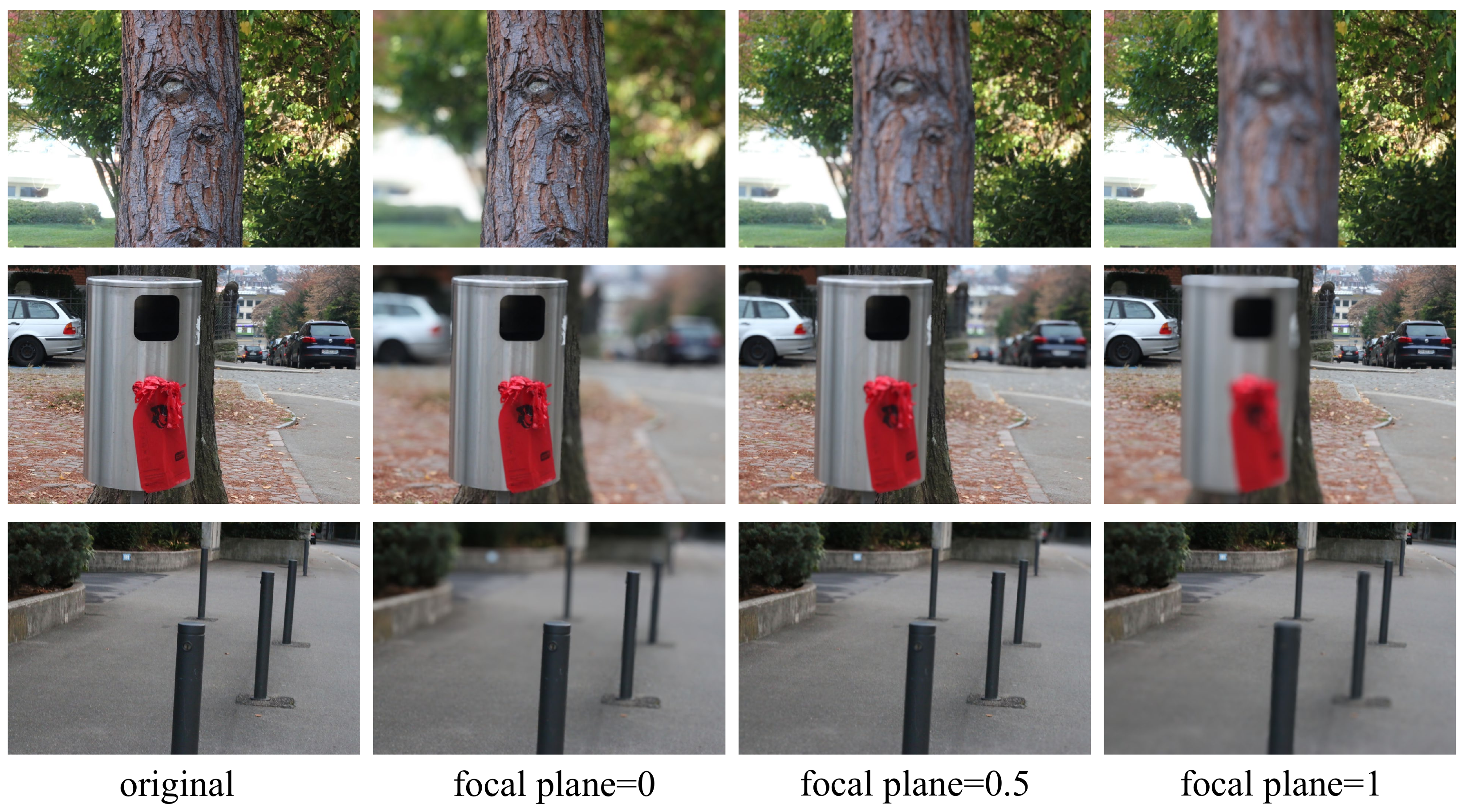}

        \caption{\textbf{Bokeh from different focal distances.} From column 2 to column 4, the focal plane is changed from foreground to background. We define the focal plane as the relative distance from the focused plane to the camera.
        }
        \label{fig:focal}
        \end{figure}
    \begin{figure}[!tbp]
        \centering
        \includegraphics[width=1.0\linewidth]{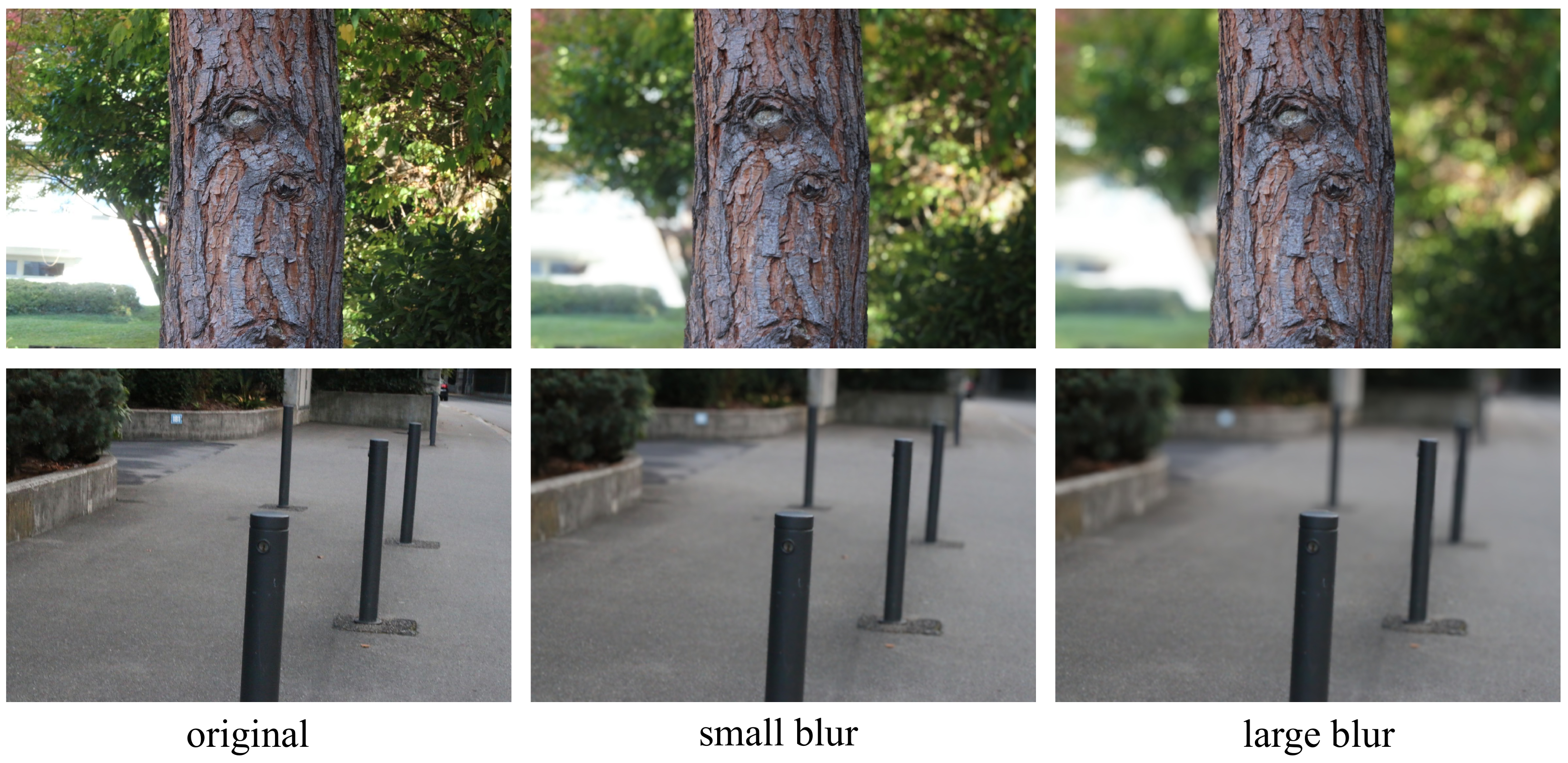}

        \caption{ Bokeh with different blur amount.}
        \label{fig:blur}
        \end{figure}

    As shown in Fig.~\ref{fig:stra}, the blue line is our current strategy that train images on 512$\times$512 and output single-channel defocus maps. In $(a)$ the green line indicates our previous three-step training method. In this scheme we train the network to produce multi-channel defocus maps on 256$\times$256 to converge faster, then we finetuned on 512$\times$512, finally, we switch to output single-channel defocus maps.
    
    The green dotted line in table $b$ is a two-step scheme that trains images on 512$\times$512, and the scheme is the same as the last two steps of the three-step strategy from table $a$. The sudden decline or rise in these two green dotted lines are resulted from changing training resolutions or the channels of output defocus maps. To achieve better CoC, we need to output single-channel defocus maps, so the last step of each scheme is the same.

    Judging from the graph, the validation loss and the PSNR result of \ourmethod are superior to those of our previous approach and another multi-stage training strategy which are originally designed to converge faster.

    \subsection{Applications}\label{subsec: applications}
    
    Since \ourmethod is trained on a dataset where foreground objects are in-focus, we can assume the focal plane is at 0 in terms of relative depth. Therefore, we can alter the defocus map in order to change the focal plane. As shown in Fig.~\ref{fig:focal}, we set the relative disparity map from 0 to 1, and we focus on three planes where the focal distance is 0, 0.5 and 1. We define the focal distance as the relative distance from the focused plane to the camera, ranged from 0 to 1. In addition, we use pre-defined blur kernels, so we can also modify the blur amount by different blur kernel sizes, as shown in Fig.~\ref{fig:blur}. It is worth mentioning that other methods from AIM 2019 Rendering Realistic Bokeh Challenge and AIM 2020 Rendering Realistic Bokeh Challenge are not able to adjust the focal plane and the blur amount.
    
    \begin{figure}[!tbp]
        \centering
        \includegraphics[width=1.0\linewidth]{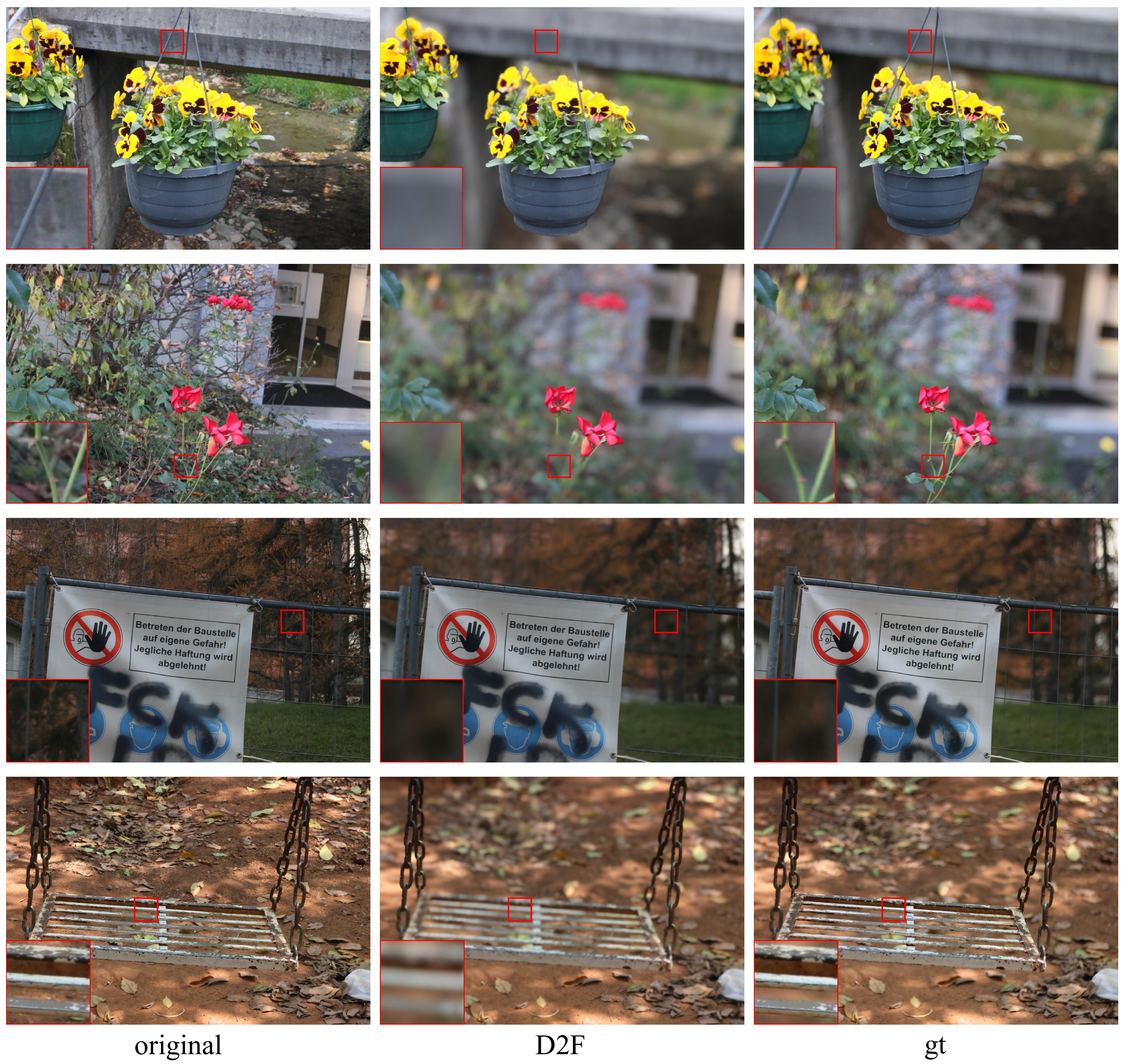}

        \caption{
        Examples of some failure cases.}
        \label{fig:failure}
        \end{figure}
	\subsection{Failure Case Analyses}\label{subsec:Failure Cases Analyses}

	Although \ourmethod achieves pleasing bokeh effects, it still has some defects. We show some failure cases in this section. As shown in Fig.~\ref{fig:failure}, some in-focus parts of the generated bokeh images are blurred by mistake. This phenomenon is caused by false defocus hallucination, and the defocus map might be inaccurate or not versatile enough because of our manual settings of blur kernels. The defocus map is likely to be inconsistent when the structure is too thin, such as the branches and fences in Fig.~\ref{fig:failure}. In addition, as shown in the row 4 from Fig.~\ref{fig:failure}, if the focused object has a stripe-like shape, it might get blurred with the surrounding background. We believe the introduction of depth map can improve the rendered result. If the depth map is accurate, it can provide with correct ordinal relationship between objects, so that \ourmethod is able to distinguish in-focus and out-of-focus regions.

	\section{Conclusion}\label{sec:conclu}
	We have presented an effective fusion framework \ourmethod to predict a bokeh image from a single narrow aperture image. By introducing defocus hallucination within our network using only a bokeh image as supervision, we train our defocus hallucination network to produce a single-channel defocus map, improving the aesthetic quality of the synthesized bokeh. 
	\ourmethod also improves the fusion of blurred images in layered rendering by radiance virtualization. 
	Radiance virtualization transforms image intensity into scene radiance, and weighted layered rendering can generate better CoC results. 
	In addition, we utilize the predicted defocus map as guidance for image fusion. Then we apply the Poisson gradient loss in our deep Poisson network to refine our initial fusion mask, which ensures the sharpness of in-focus parts. 
	Exhaustive visualizations and ablation studies are presented to validate these modules and demonstrate their effects on the performance of \ourmethod. We significantly enhance the quality of bokeh images. In the future, we can explore using dynamic filters for the selection of kernels and modify the defocus hallucination module to learn a more accurate single-channel defocus map.





\section*{Acknowledgement}
{This work was supported in part by the National Natural Science Foundation of China under Grant No. U1913602.
			
			}


\bibliographystyle{elsarticle-num}
\bibliography{refs}


\end{document}